\documentclass[10pt,twocolumn,letterpaper]{article}
\usepackage[pagenumbers]{cvpr}
\usepackage{multirow}
\usepackage{color}
\usepackage{amsmath}
\usepackage{bm}

\usepackage{placeins}
\usepackage{float}
\usepackage{stfloats}
\definecolor{red}{rgb}{1,0.2,0.2}
\definecolor{or}{rgb}{1,0.5,0.25}
\definecolor{green}{rgb}{0, 1, 0}
\definecolor{bl}{rgb}{0, 0, 1}
\definecolor{brown}{rgb}{0.59, 0.3, 0}
\definecolor{cyan}{rgb}{0, 1, 1}
\definecolor{c_lowbest}{rgb}{1.0,1.0,0.9}
\definecolor{c_highbest}{rgb}{0.9,1.0,0.9}

\newcommand{\best}[1]  {\textcolor{red}{\textbf{#1}}}
\newcommand{\second}[1]  {\textcolor{blue}{\underline{#1}}}

\newcommand*{\affaddr}[1]{#1} 
\newcommand*{\affmark}[1][*]{\textsuperscript{#1}}
\newcommand*{\email}[1]{\texttt{#1}} %

\definecolor{cvprblue}{rgb}{0.21,0.49,0.74}
\usepackage[pagebackref,breaklinks,colorlinks,allcolors=cvprblue]{hyperref}

\title{\includegraphics[scale=0.4, keepaspectratio]{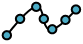}SplineGS: Robust Motion-Adaptive Spline \\ for Real-Time Dynamic 3D Gaussians from Monocular Video}

\author{
Jongmin Park\affmark[1]\footnotemark[1] \quad Minh-Quan Viet Bui\affmark[1]\footnotemark[1]\quad Juan Luis Gonzalez Bello\affmark[1]\quad Jaeho Moon\affmark[1] \\ Jihyong Oh\affmark[2]\footnotemark[2] \quad Munchurl Kim\affmark[1]\footnotemark[2]\\
\affaddr{\affmark[1]KAIST} \quad
\affaddr{\affmark[2]Chung-Ang University}\\
\small{\email{\{jm.park, bvmquan, juanluisgb, jaeho.moon, mkimee\}@kaist.ac.kr}} \quad
\small{\email{jihyongoh@cau.ac.kr}}\\
\affaddr{\small{\url{https://kaist-viclab.github.io/splinegs-site/}}}%
}

\begin{document}

\twocolumn[{
\renewcommand\twocolumn[1][]{#1}%
\maketitle
    \begin{center}
           \includegraphics[width=\linewidth,keepaspectratio]{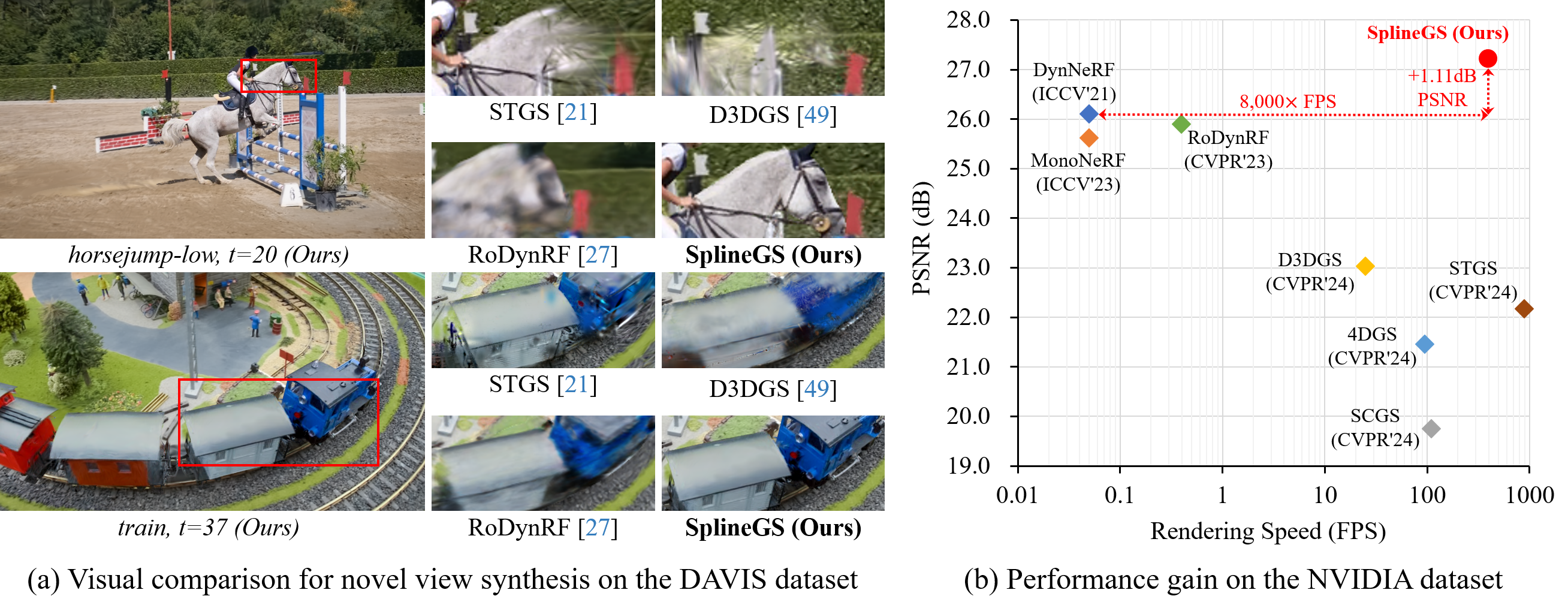}
           \vspace{-0.8cm}
           \captionof{figure}{\textbf{Our SplineGS achieves state-of-the-art rendering quality with fast rendering speed for novel spatio-temporal view synthesis from monocular videos without relying on pre-computed camera parameters.} (a) We use our predicted camera parameters for \cite{yang2023deformable3dgs, Li_STG_2024_CVPR} since COLMAP~\cite{schonberger2016structure} is unable to provide reasonable camera parameters for most scenes in the DAVIS dataset~\cite{ponttuset20182017davischallengevideo}. (b) SplineGS achieves 1.1 dB higher PSNR and 8,000$\times$ faster rendering speed compared to the second-best method on the NVIDIA dataset~\cite{yoon2020dynamic}.}
           \label{fig:figure_page1}
    \end{center}
}]

\maketitle

\begin{abstract}
Synthesizing novel views from in-the-wild monocular videos is challenging due to scene dynamics and the lack of multi-view cues. To address this, we propose SplineGS, a COLMAP-free dynamic 3D Gaussian Splatting (3DGS) framework for high-quality reconstruction and fast rendering from monocular videos. At its core is a novel Motion-Adaptive Spline (MAS) method, which represents continuous dynamic 3D Gaussian trajectories using cubic Hermite splines with a small number of control points. For MAS, we introduce a Motion-Adaptive Control points Pruning (MACP) method to model the deformation of each dynamic 3D Gaussian across varying motions, progressively pruning control points while maintaining dynamic modeling integrity. Additionally, we present a joint optimization strategy for camera parameter estimation and 3D Gaussian attributes, leveraging photometric and geometric consistency. This eliminates the need for Structure-from-Motion preprocessing and enhances SplineGS’s robustness in real-world conditions. Experiments show that SplineGS significantly outperforms state-of-the-art methods in novel view synthesis quality for dynamic scenes from monocular videos, achieving \textbf{thousands} times faster rendering speed.
\vspace{-0.5cm}
\end{abstract}

{
  \renewcommand{\thefootnote}%
    {\fnsymbol{footnote}}
  \footnotetext[1]{Co-first authors (equal contribution).}
  \footnotetext[2]{Co-corresponding authors.}
}

\section{Introduction}
\label{sec:intro}

Novel View Synthesis (NVS) is fundamental to 3D vision, supporting applications like virtual reality (VR), augmented reality (AR), and film production. NVS aims to generate images from any viewpoint in a scene, requiring accurate reconstruction from multiple 2D images. NeRF \cite{mildenhall2020nerf} has advanced the field of NVS by utilizing learned implicit functions to represent static scenes, though it requires considerable training and rendering time. Recently, 3D Gaussian Splatting (3DGS) \cite{kerbl20233d} has revolutionized this process by replacing implicit volumetric rendering with differentiable rasterization of 3D Gaussians, enabling real-time rendering and providing a more explicit scene representation.


For NVS of dynamic scenes, prior works have extended the static scene representations in \cite{kerbl20233d, mildenhall2020nerf} by incorporating deformation models in canonical space using implicit representations \cite{park2021hypernerf, liu2023robust, yang2023deformable3dgs, Gao-ICCV-DynNeRF, park2021nerfies}, grid-based models that decompose the 4D space-time domain into multiple 2D planes \cite{fridovich2023k, cao2023hexplane, Wu_2024_CVPR, attal2023hyperreel, shao2023tensor4d}, and polynomial trajectories \cite{Li_STG_2024_CVPR}. However, these methods face several challenges: implicit representations significantly increase computational overhead and reduce rendering speed \cite{cao2023hexplane, Wu_2024_CVPR}; grid-based models struggle to fully capture the dynamic nature of scene structures, hindering their ability to model fine details accurately \cite{Li_STG_2024_CVPR}; and although polynomial trajectories improve efficiency, their fixed degree restricts flexibility for representing complex motions. To the best of our knowledge, none of the dynamic 3DGS-based methods provide experimental evidence to reliably support their novel spatio-temporal view synthesis capabilities for rendering unseen intermediate time indices. Additionally, most existing methods for dynamic NVS rely heavily on external camera parameter estimation methods like COLMAP \cite{schonberger2016structure}, which often produce imprecise results for challenging in-the-wild monocular videos.

To address the aforementioned issues for modeling scene dynamics, we exploit a spline-based model to dynamic 3D Gaussian trajectories, inspired by classic 3D-curve modeling in computer graphics~\cite{farin2001curves}. Splines are widely used in geometric modeling for graphical applications, providing smooth and continuous representations of complex shapes with a minimal number of control points~\cite{ahlberg2016theory, de1978practical}. This efficiency makes them ideal for modeling intricate geometric structures while maintaining both flexibility and precision. In this paper, we propose SplineGS, a framework for high-quality dynamic scene reconstruction and real-time neural rendering from \textit{in-the-wild} monocular videos without relying on external camera estimators like COLMAP \cite{schonberger2016structure}. SplineGS introduces Motion-Adaptive Spline (MAS), based on cubic Hermite splines~\cite{ahlberg2016theory, de1978practical}, to effectively represent dynamic 3D Gaussian deformations. Our MAS consists of piecewise cubic functions defined by control points that dictate each segment’s curvature and direction. Each control point is adjustable and optimized as a learnable parameter, enabling faster and more precise modeling of complex motion trajectories. Additionally, to adaptively model the trajectory of each dynamic 3D Gaussian based on motion complexity during training, we introduce a Motion-Adaptive Control points Pruning (MACP) method that adjusts the number of control points to improve rendering quality and efficiency. Furthermore, we incorporate a camera parameter estimation method jointly optimized with 3D Gaussian attributes and MAS, leveraging photometric and geometric consistency, thus eliminating the need for external estimators. Experiments show that SplineGS \textit{significantly outperforms} state-of-the-art (SOTA) dynamic NVS methods both qualitatively and quantitatively, offering faster rendering speed, as shown in Fig.~\ref{fig:figure_page1}. 
Our contributions are as follows:

\begin{itemize}
 \setlength\itemsep{0.1cm}
  \item We propose a novel \textit{\textbf{Spline}-based} dynamic 3D \textbf{G}aussian \textbf{S}platting framework, SplineGS, which is (i) \textit{COLMAP-free}, (ii) \textit{very fast} and (iii) \textit{of high quality} in reconstructing the dynamic scenes from in-the-wild monocular videos;
  \item A novel \textbf{M}otion-\textbf{A}daptive \textbf{S}pline (MAS) is introduced, which can accurately and effectively represent the continuous trajectory of each dynamic 3D Gaussian;
  \item A \textbf{M}otion-\textbf{A}daptive \textbf{C}ontrol points \textbf{P}runing (MACP) method is presented, which can efficiently adjust the number of control points for each spline function, optimizing both rendering quality and the efficiency of MAS.
\end{itemize}
\section{Related Work}
\label{sec:related_work}
\noindent \textbf{Dynamic NeRF.} Recent advancements in video view synthesis have extended the static NeRF model \cite{mildenhall2020nerf} to represent scene dynamics. These include dynamic NeRFs using scene flow-based frameworks \cite{Gao-ICCV-DynNeRF, li2021neural, li2023dynibar}, deformation estimation in canonical fields \cite{park2021nerfies, park2021hypernerf, pumarola2021d, tretschk2021non, weng_humannerf_2022_cvpr, jiang2022neuman, yang2022banmo, athar2022rignerf, fang2022fast, song2023nerfplayer, liu2023robust}, and 4D grid-based spatio-temporal radiance fields \cite{cao2023hexplane, fridovich2023k, shao2023tensor4d, attal2023hyperreel}. Techniques such as NSFF \cite{li2021neural}, DynNeRF \cite{Gao-ICCV-DynNeRF}, and DynIBaR \cite{li2023dynibar} combine time-independent and time-dependent radiance fields to synthesize novel spatio-temporal perspectives from monocular videos. Despite these advancements, current dynamic NeRFs still fall short of recent dynamic 3D Gaussian Splatting (3DGS) techniques in rendering quality and efficiency.

\noindent \textbf{Dynamic 3DGS.} The improvements in rendering quality and speed achieved by 3DGS \cite{kerbl20233d} have inspired further studies \cite{luiten2023dynamic, liang2023gaufre, yang2023deformable3dgs, Wu_2024_CVPR, Li_STG_2024_CVPR, huang2023sc} to extend the static 3DGS framework to dynamic scenes by enabling the deformation of 3D Gaussian attributes. The pioneering work on dynamic 3DGS \cite{luiten2023dynamic} introduces time-dependent offsets for the positions and rotations of dynamic 3D Gaussians via an MLP; however, this approach slows down the rendering. D3DGS \cite{yang2023deformable3dgs} builds on this concept with an annealing smoothing training mechanism to improve temporal smoothness and rendering quality. 4DGS \cite{Wu_2024_CVPR} replaces the MLP network with a grid-based structure to boost efficiency, though this change requires quality trade-offs due to the resolution limits of grid-based methods. In contrast, SC-GS \cite{huang2023sc} combines an MLP deformation network with sparse spatial deformation, reducing the computational cost of MLP while maintaining quality. STGS \cite{Li_STG_2024_CVPR} introduces polynomial trajectories for motion modeling, improving speed and quality over implicit representations. Casual-FVS \cite{lee2023casual-fvs} warps dynamic content from neighboring frames, and MoSca \cite{lei2024moscadynamicgaussianfusion} proposes a graph-based motion modeling approach to handle sparse control deformation. Very recently, a concurrent work~\cite{lee2024fully} proposes modeling dynamic 3D Gaussian trajectories using polynomial interpolation between \textit{fixed} keyframes for NVS from \textit{multi-view} videos \textit{with COLMAP~\cite{schonberger2016structure} assistance}. Unlike ~\cite{lee2024fully}, our SplineGS \textit{adaptively} optimizes the deformation of dynamic 3D Gaussians, accounting for varying motion degrees and types in in-the-wild \textit{monocular} videos, without requiring preprocessed camera parameters (\textit{COLMAP-free} approach).

\noindent \textbf{Neural Rendering without SfM Preprocessing.}
Accurate camera parameters, including extrinsics and intrinsics, are essential for neural rendering approaches to capture fine details \cite{jiang2023alignerf, lin2021barf}. However, in real-world settings, camera parameters derived from Structure-from-Motion (SfM) algorithms such as COLMAP~\cite{schonberger2016structure} often exhibit pixel-level inaccuracies \cite{raoult2017reliable, lindenberger2021pixel}, compromising the structural details of rendered scenes \cite{liu2023robust}. To address this, several NeRF methods \cite{lin2021barf, nerfmm, camp, meng2021gnerf} jointly optimize NeRF architectures and camera parameters. Recently, a local-to-global training approach \cite{Fu_2024_CVPR} is introduced to optimize both camera parameters and 3D Gaussians. However, these methods are limited to static scenes. For dynamic scene reconstruction, RoDynRF \cite{liu2023robust} and MoSca \cite{lei2024moscadynamicgaussianfusion} use motion masks to gather multi-view cues from static regions, allowing robust rendering without pre-computed camera parameters. Our SplineGS is also COLMAP-free, significantly outperforming RoDynRF \cite{liu2023robust} and MoSca \cite{lei2024moscadynamicgaussianfusion} in rendering quality and efficiency, enabled by our novel spline-based architecture.
\section{Preliminary: 3D Gaussian Splatting}
\label{sec:preliminary}

3DGS~\cite{kerbl20233d} represents the radiance field of a scene using anisotropic 3D Gaussians, each of which is formulated as
\vspace{-0.1cm}
\begin{equation}
    G(\bm{x}) = \text{exp}({-(1/2) (\bm{x}-\bm{\mu})^\top \bm{\Sigma}^{-1} (\bm{x}-\bm{\mu}))},
    \vspace{-0.1cm}
\label{eq:Gaussians}
\end{equation}
where $\bm{x} \in \mathbb{R}^3$ denotes a 3D position, and $\bm{\mu} \in \mathbb{R}^3$ and $\bm{\Sigma} \in \mathbb{R}^{3\times3}$ represent the mean (center) and the covariance matrix of the 3D Gaussian, respectively. 
To ensure that $\bm{\Sigma}$ is positive semi-definite and contains physical meaning, it is decomposed into a diagonal scaling matrix $\bm{S} \in \mathbb{R}^{3\times3}$ of a scale vector $\bm{s} \in \mathbb{R}^3
$ and a rotation matrix $\bm{R} \in \mathbb{R}^{3\times3}$ as $\bm{\Sigma} = \bm{R}\bm{S}\bm{S}^{\top}\bm{R}^{\top}$, where $\bm{R}$ is parameterized by a learnable unit-length quaternion $\bm{q} \in \mathbb{R}^4$. In addition, each 3D Gaussian is parameterized by an opacity $\sigma \in \mathbb{R}$ and a color $\bm{c} \in \mathbb{R}^3$. To render the color of each pixel, the color and the opacity of each 3D Gaussian are computed using Eq.~\ref{eq:Gaussians}, and the rendered color $\bm{C}$ is computed by the alpha-blending of the $\mathcal{N}$ ordered 3D Gaussians overlapping the pixel as
\vspace{-0.1cm}
\begin{equation}
    \bm{C} = \textstyle \sum_{i\in\mathcal{N}}\bm{c}_i\alpha_i\prod^{i-1}_{j=1}(1-\alpha_j),
    \vspace{-0.1cm}
    \label{eq:alpha_blending}
\end{equation}
where $\bm{c}_i$ is the color of the $i^{\text{th}}$ 3D Gaussian and $\alpha_i$ is a density of the $i^{\text{th}}$ 3D Gaussian which is given by evaluating the 2D covariance $\bm{\Sigma}' \in \mathbb{R}^{2\times2}$. Here, $\bm{\Sigma}'$ is formulated as $\bm{\Sigma}' = \bm{JW\Sigma W^{\top}J^{\top}}$, where $\bm{J}$ is the Jacobian of the affine approximation of the projective transformation and $\bm{W}$ is a viewing transformation matrix.

Similar to prior works \cite{liang2023gaufre, lei2024moscadynamicgaussianfusion}, we extend 3DGS~\cite{kerbl20233d} to a union of static 3D Gaussians $\{G^\text{st}_i|i=1, 2, ..., n^{\text{st}}\}$ and dynamic 3D Gaussians $\{G^\text{dy}_i|i=1, 2, ..., n^{\text{dy}}\}$ to represent static backgrounds and moving objects, respectively, in dynamic scenes. We maintain the same gradient-based densification \cite{kerbl20233d} for both static $\{G^\text{st}_i\}$ and dynamic $\{G^\text{dy}_i\}$ 3D Gaussians. Following STGS \cite{Li_STG_2024_CVPR}, we use the splatted feature rendering to predict the final pixel colors. For static regions, we remove the time-encoded feature while preserving the diffuse and specular features. We model the mean $\bm{\mu}_i$ of $G^\text{dy}_i$ as a time-dependent variable, defined by our novel deformation modeling method. We compute the time-dependent rotation $\bm{q}_i$ and scale $\bm{s}_i$ by modeling them as learnable parameters of time. 

\section{Proposed Method: SplineGS}
\label{sec:proposed_method}

\begin{figure*}
\centering
\includegraphics[width=\linewidth,keepaspectratio]{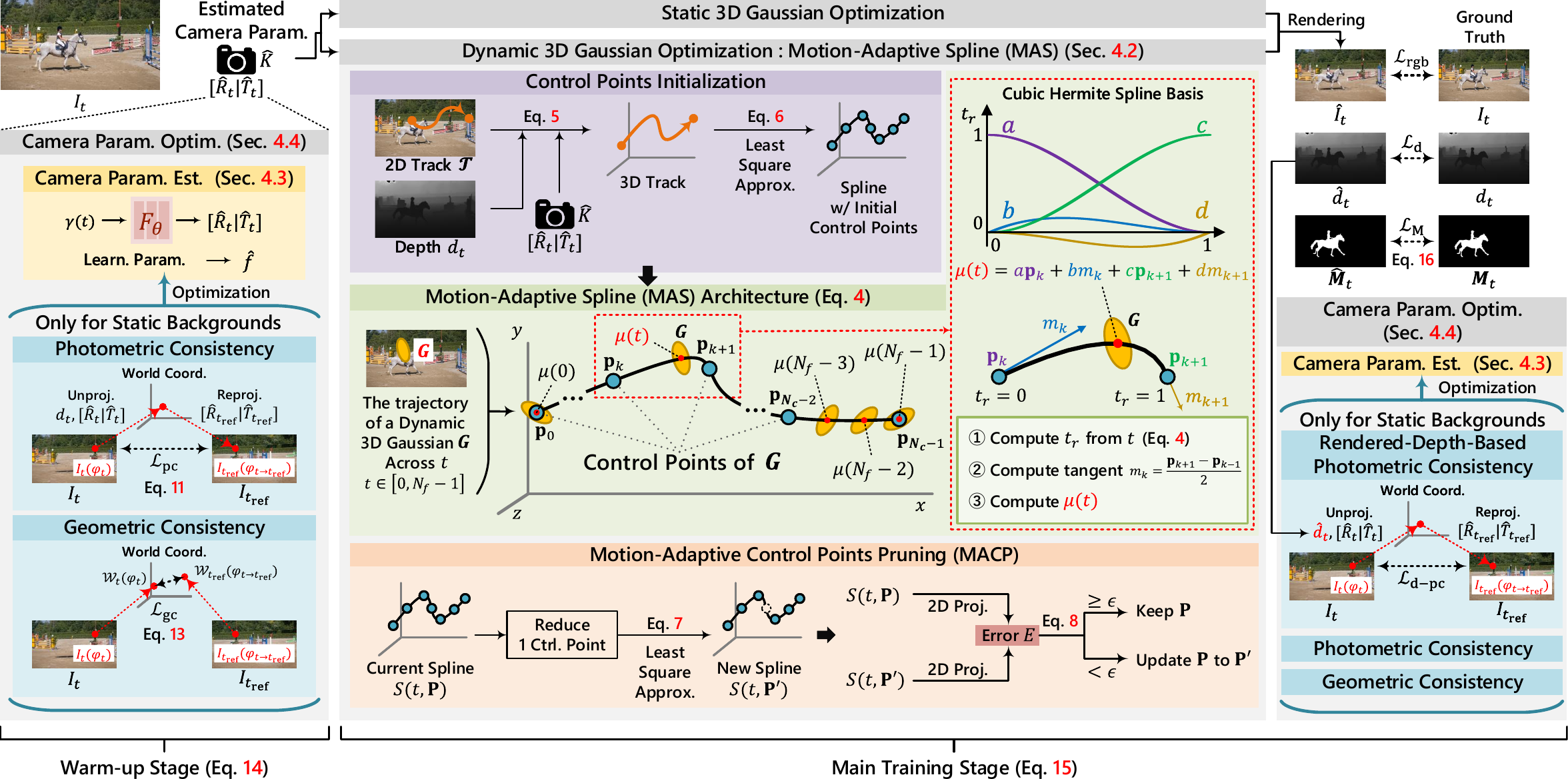}
\vspace{-0.5cm}
\caption{\textbf{Overview of SplineGS.} Our SplineGS leverages spline-based functions to model the deformation of dynamic 3D Gaussians with a novel Motion-Adaptive Spline (MAS) architecture. It is composed of sets of learnable control points based on a cubic Hermite spline function~\cite{ahlberg2016theory, de1978practical} to accurately model the trajectory of each dynamic 3D Gaussian and to achieve faster rendering speed. To avoid any preprocessing of camera parameters, i.e. COLMAP-free, we adopt a two-stage optimization: warm-up and main training stages.}
\label{fig:overall_architecture}
\vspace{-0.5cm}
\end{figure*}

\subsection{Overview of SplineGS}
We describe the overall architecture of SplineGS in Fig.~\ref{fig:overall_architecture}. Given a monocular video $\{\bm{I}_t|t=0,1,...,N_f-1\}$ where $N_f$ is the total number of frames, SplineGS is designed to synthesize high-quality novel spatio-temporal views with fast rendering speed, and to estimate the camera parameters, including the extrinsics $[\hat{\bm{R}}_t|\hat{\bm{T}}_t] \in \mathbb{R}^{3\times4}$ for each time $t$, and the shared intrinsic $\hat{\bm{K}} \in \mathbb{R}^{3\times3}$.
As shown in Fig.~\ref{fig:overall_architecture}, to stabilize joint optimization of the 3D Gaussian attributes and camera parameter estimation, we adopt a two-stage optimization process consisting of a warm-up stage and a main training stage for our SplineGS architecture. In the warm-up stage, we coarsely optimize the camera parameters $[\hat{\bm{R}}_t|\hat{\bm{T}}_t]$ and $\hat{\bm{K}}$ using photometric and geometric consistency. In the main training stage, we initialize the 3D Gaussians based on the estimated camera poses and jointly optimize the 3D Gaussian attributes with the camera parameter estimation. Specifically, for each dynamic 3D Gaussian, we propose a novel spline-based deformation modeling method, called Motion-Adaptive Spline (MAS), which utilizes a cubic Hermite spline function~\cite{ahlberg2016theory, de1978practical} to accurately model the continuous trajectory with a small number of control points. For MAS, we introduce a Motion-Adaptive Control points Pruning (MACP) method that effectively adjusts the number of control points for each dynamic 3D Gaussian, taking into account the object's motion types and degrees.

In the following sections, we detail the process of our MAS method in Sec.~\ref{subsec:splineGS}, followed by the camera parameter estimation process in Sec.~\ref{subsec:camera_pose_optim}. Finally, we describe the overall optimization process in Sec.~\ref{subsec:optim}.

\subsection{Motion-Adaptive Spline for 3D Gaussians}
\label{subsec:splineGS}
To represent the continuous trajectory of \textit{each} dynamic 3D Gaussian for moving objects over time, we propose MAS which modifies the mean parameter to a set of learnable control points. For this, we use the cubic Hermite spline function~\cite{ahlberg2016theory, de1978practical} to model the mean of each \textit{dynamic} 3D Gaussian at time $t$ as
\vspace{-0.1cm}
\begin{equation}
    \mu(t) = S(t, \textbf{P}),
    \vspace{-0.1cm}
\end{equation}
where $\textbf{P} = \{\textbf{p}_k | \textbf{p}_k \in \mathbb{R}^3 \}_{k \in [0, N_c-1]}$ is a set of $N_c$ learnable control points, serving as an additional attribute for each dynamic 3D Gaussian, and $S(t, \textbf{P})$ is formulated as
\vspace{-0.2cm}
\begin{equation}
\scalebox{0.95}{$
\begin{aligned}
    S(t, \textbf{P}) = (2t_r^3 - 3t_r^2 + 1)\textbf{p}_{\lfloor t_s \rfloor} + (t_r^3 - 2t_r^2 + t_r)\textbf{m}_{\lfloor t_s \rfloor} \\
    + (-2t_r^3 + 3t_r^2)\textbf{p}_{\lfloor t_s \rfloor +1} + (t_r^3 - t_r^2)\textbf{m}_{\lfloor t_s \rfloor +1},  \\
    t_r = t_s - \lfloor t_s \rfloor, \quad t_s = t_n(N_c-1), \quad t_n = t / (N_f-1),
\end{aligned}
$}
\label{eq:spline}
\vspace{-0.2cm}
\end{equation}
where  $\textbf{m}_k$ is the approximated tangent of the control point $\textbf{p}_k$, formulated as $\textbf{m}_k = (\textbf{p}_{k+1} - \textbf{p}_{k-1})/2$. Note that $S(t, \textbf{P})$ indicates the piecewise cubic function to represent the whole continuous trajectory of each dynamic 3D Gaussian, and $N_c$ can be different from $N_f$. The optimal $N_c$ can be estimated by Motion-Adaptive Control points Pruning (MACP) in the following section.

\noindent \textbf{Control Points Initialization.} To stably optimize $S(t, \textbf{P})$, it is essential to initialize the control points with appropriate geometric considerations that ensure temporal consistency. For this, we leverage the long-range 2D tracking~\cite{karaev2023cotracker} and the metric depth~\cite{piccinelli2024unidepth} priors. Let $\bm{\mathcal{T}} = \{\bm{\varphi}^\text{tr}_t | \bm{\varphi}^\text{tr}_t \in \mathbb{R}^2 \}_{t \in [0, N_f-1]}$ be a 2D track, $\pi_{\bm{K}}(\cdot)$ be a projection function from the camera space to the image space with the camera intrinsic $\bm{K}$, and $\bm{\varphi}^\text{tr}_t$ be a pixel coordinate corresponding to the 2D track $\bm{\mathcal{T}}$ at time $t$. We unproject each 2D track $\bm{\mathcal{T}}$ to the world space to compute the 3D track aided by the frame's metric depth $\bm{d}_t$ and camera extrinsic $[\hat{\bm{R}}_t|{\hat{\bm{T}}}_t]$ at time $t$. The unprojection function $\mathcal{W}_{t}(\cdot)$ from the image space to the world space is formulated as
\vspace{-0.2cm}
\begin{equation}
    \mathcal{W}_{t}(\bm{\varphi}^\text{tr}_t) = \hat{\bm{R}}_{t}^\top\pi_{\hat{\bm{K}}}^{-1}\big(\bm{\varphi}^\text{tr}_t, \bm{d}_{t}(\bm{\varphi}^\text{tr}_t)\big) - \hat{\bm{R}}_{t}^\top{\hat{\bm{T}}_{t}}.
\label{eq:unproj}
\vspace{-0.2cm}
\end{equation}
It should be noted that we estimate $[\hat{\bm{R}}_{t}|\hat{\bm{T}}_{t}]$ and $\hat{\bm{K}}$ from a sequence of frames only, without using any given ground truth values, as mentioned in Sec.~\ref{subsec:camera_pose_optim}. Then, we initialize the per-Gaussian control points set $\textbf{P}$, by a least-square (LS) approximation, such that the spline-described curve fits the initial curve by the tracker, given by
\vspace{-0.2cm}
\begin{equation}
    \min_{{\textbf{P}}} \textstyle \sum^{N_f-1}_{t=0} \big\|\mathcal{W}_{t}(\bm{\varphi}^\text{tr}_t) - S(t, \textbf{P}) \big\|^2_2.
    \vspace{-0.2cm}
\end{equation}

\noindent \textbf{Motion-Adaptive Control Points Pruning (MACP).} An excessive number of control points may cause over-fitting and decrease the processing speed of our MAS module, leading to poorer rendering qualities with reduced rendering speed (see Table~\ref{table:ablation_study}).
Furthermore, as each scene exhibits various types and degrees of motion for moving objects, the number of control points required for each dynamic 3D Gaussian trajectory should be adaptively adjusted to accommodate the scene dynamics. To achieve this, we propose the MACP method for MAS, which can generate sparser control points while ensuring no dynamic modeling degradation. MACP is designed on top of 3D Gaussian densification \cite{kerbl20233d}, but focuses on optimizing the number of control points in MAS. Our MACP computes a new spline function $S(t, \textbf{P}')$ after every 3D Gaussian densification step, where $\textbf{P}' = \{\textbf{p}'_l | \textbf{p}'_l \in \mathbb{R}^3 \}_{l \in [0, N_c-2]}$ is a set of $N_c-1$ control points, which contains one-fewer control points than the current set $\textbf{P}$ of control points. We compute the LS approximation to find the reduced set $\textbf{P}'$ of control points as
\vspace{-0.2cm}
\begin{equation}
    \min_{{\textbf{P}'}} \textstyle \sum^{N_f-1}_{t=0} \big\|S(t, \textbf{P}) - S(t, \textbf{P}') \big\|^2_2.
    \vspace{-0.2cm}
\end{equation}
Then, the updated optimal set $\textbf{P}$ of control points is assigned with the values of $\textbf{P}'$ if the error $E$ between $S(t, \textbf{P})$ and $S(t, \textbf{P}')$ is smaller than a threshold $\epsilon$, as given by 
\vspace{-0.2cm}
\begin{equation}
\scalebox{0.84}{$
    \begin{aligned}
          &\textbf{P}=\begin{cases}
        			\textbf{P}', ~ \text{if $E < \epsilon$}\\
                    \textbf{P}, ~ \text{otherwise}
        		\end{cases}, ~ \text{where} \\ 
          \hspace{-0.69em}E = \frac{1}{N_f}
          \textstyle \sum_{t=0}^{N_f-1} \|&\pi_{\hat{\bm{K}}}(\hat{\bm{R}}_{t}S(t, \textbf{P}) + \hat{\bm{T}}_{t}) - \pi_{\hat{\bm{K}}}(\hat{\bm{R}}_{t}S(t, \textbf{P}') + \hat{\bm{T}}_{t})\|^2_2.
    \end{aligned}
$}
\label{eq:2d_thershold_macp}
\end{equation}
By following MACP, each dynamic 3D Gaussian is allowed to have a different number of control points. Therefore, the MACP can guide our MAS to have a minimal number of control points when modeling simple motion, while having more control points for more complex motions (see Fig. \ref{fig:macp}).

\subsection{Camera Parameter Estimation}
\label{subsec:camera_pose_optim}
The traditional SfM methods, such as COLMAP~\cite{schonberger2016structure}, fail to reliably estimate camera parameters in dynamic scenes from in-the-wild monocular videos \cite{liu2023robust}. For this reason, we propose to estimate camera parameters for joint optimization with the 3D Gaussian attributes. For each frame at time $t$, we predict a rotation matrix $\hat{\bm{R}}_t \in \mathbb{R}^{3\times3}$ and a translation vector $\hat{\bm{T}}_t \in \mathbb{R}^{3}$, representing the extrinsic parameters of a monocular camera, using a shallow MLP $F_\theta$ as
\vspace{-0.1cm}
\begin{equation}
    [\hat{\bm{R}}_t|\hat{\bm{T}}_t] = F_\theta(\gamma(t)),
\label{eq:cam_prediction}
\vspace{-0.1cm}
\end{equation}
where $\gamma(\cdot)$ is a positional encoding~\cite{mildenhall2020nerf}. We also predict the focal length $\hat{f}$ of the camera intrinsic $\hat{\bm{K}}$ as a learnable parameter that is shared across all frames in the monocular video. To accurately optimize the camera parameters, we enforce two types of consistency—photometric and geometric—for the static background between random reference cameras $[\hat{\bm{R}}_{t_\text{ref}}|\hat{\bm{T}}_{t_\text{ref}}]$ and the target camera $[\hat{\bm{R}}_t|\hat{\bm{T}}_t]$, encouraging alignment both visually and structurally.

\noindent \textbf{Photometric Consistency.} 
Given the pre-computed metric depth \cite{piccinelli2024unidepth}, the camera intrinsics and extrinsics under optimization will converge as long as the projected reference frame's color $\bm{I}_{t_\text{ref}}(\bm{\varphi}_{t \rightarrow t_\text{ref}})$ is aligned to the target frame's color $\bm{I}_t(\bm{\varphi}_t)$. $\bm{\varphi}_{t \rightarrow t_\text{ref}}$ is the reference pixel coordinate corresponding to the target frame's pixel coordinate $\bm{\varphi}_t$ as
\vspace{-0.2cm}
\begin{equation}
\scalebox{0.85}{$
    \bm{\varphi}_{t \rightarrow t_\text{ref}} = \pi_{\hat{\bm{K}}}(\hat{\bm{R}}_{t_\text{ref}}(\hat{\bm{R}}_{t}^\top\pi_{\hat{\bm{K}}}^{-1}\big(\bm{\varphi}_t, \bm{d}_{t}(\bm{\varphi}_t)\big) - \hat{\bm{R}}_{t}^\top{\hat{\bm{T}}_{t}}) + \hat{\bm{T}}_{t_\text{ref}}).
$}
\label{eq:unproj_pc}
\vspace{-0.2cm}
\end{equation}
We refer to such projection alignment as photometric consistency, which is encouraged by the loss $\mathcal{L}_{\text{pc}}$ given by 
\begin{equation}
\scalebox{0.9}{$
     \mathcal{L}_{\text{pc}} = \textstyle \sum_{\bm{\varphi}_t} \big\| \bm{M}_{t,t_\text{ref}}(\bm{\varphi}_t) \odot (\bm{I}_t(\bm{\varphi}_t) - \bm{I}_{{t_\text{ref}}}( \bm{\varphi}_{t \rightarrow t_\text{ref}}))\big\|^2_2,
$}
\end{equation}
where $\odot$ is the Hadamard product~\cite{horn2012matrix}, and $\bm{M}_{t, t_\text{ref}}$ is a union motion mask that excludes dynamic objects in $\bm{I}_t$ and $\bm{I}_{t_\text{ref}}$ for removing the inconsistencies due to dynamic regions, which is given by
\begin{equation}
    \bm{M}_{t,t_\text{ref}}(\bm{\varphi}_t) = \bm{M}_t(\bm{\varphi}_t) \bm{M}_{t_\text{ref}}(\bm{\varphi}_{t \rightarrow t_\text{ref}}),
\label{eq:warping}
\end{equation}
where $\bm{M}_t$ and $\bm{M}_{t_\text{ref}}$ are pre-computed motion masks~\cite{yang2023track} of $\bm{I}_t$ and $\bm{I}_{t_\text{ref}}$, respectively. Note that motion mask's value is 0 for static regions, and 1 otherwise.

\vspace{1mm}
\noindent \textbf{Geometric Consistency.} Along with the photometric consistency, we compute the geometric consistency of unprojected pixels in 3D space to make our optimization more stable. The geometric consistency loss is formulated as
\begin{equation}
\scalebox{0.9}{$
\begin{aligned}
    \mathcal{L}_{\text{gc}} = \textstyle \sum_{\bm{\varphi}_t} \big\| \bm{M}_{t,t_\text{ref}}(\bm{\varphi}_t) \odot (\mathcal{W}_{t}(\bm{\varphi}_t) - \mathcal{W}_{t_\text{ref}}(\bm{\varphi}_{t \rightarrow t_\text{ref}}))\big\|^2_2,
\end{aligned}
$}
\end{equation}
where $\mathcal{W}_{t}(\bm{\varphi}_t)$ and $\mathcal{W}_{t_\text{ref}}(\bm{\varphi}_{t \rightarrow t_\text{ref}})$ are the corresponding 3D locations of $\bm{\varphi}_t$ and $\bm{\varphi}_{t \rightarrow t_\text{ref}}$, respectively (see Eq. \ref{eq:unproj}).

\subsection{Optimization}
\label{subsec:optim}
To stabilize joint training of the MAS and the camera parameter estimation, we adopt a two-stage optimization process consisting of the warm-up stage and the main training stage. In the warm-up stage, we optimize only the camera parameters using the photometric and geometric consistency. The total loss in the warm-up stage is given by
\begin{equation}
    \mathcal{L}_{\text{total}}^{\text{warm}} = \lambda_{\text{pc}} \mathcal{L}_{\text{pc}} + \lambda_{\text{gc}} \mathcal{L}_{\text{gc}}.
\end{equation}
After the warm-up stage, we obtain the coarsely predicted camera intrinsic $\hat{\bm{K}}$ and the set of extrinsics $[\hat{\bm{R}}|\hat{\bm{T}}]$ for all frames, which are then used to initialize the set $\textbf{P}$ of control points for each dynamic 3D Gaussian, as described in Sec.~\ref{subsec:splineGS}. In the main training stage, we jointly optimize the static and dynamic 3D Gaussians along with the camera parameter estimation based on the total loss function as
\begin{equation}
\begin{aligned}
    \mathcal{L}^{\text{main}}_{\text{total}} &= \lambda_{\text{rgb}} \mathcal{L}_{\text{rgb}} + \lambda_{\text{d}} \mathcal{L}_{\text{d}} + \lambda_{\text{M}} \mathcal{L}_{\text{M}} \\ 
    &+ \lambda_{\text{pc}} \mathcal{L}_{\text{pc}}
    + \lambda_{\text{d-pc}} \mathcal{L}_{\text{d-pc}}
    + \lambda_{\text{gc}} \mathcal{L}_{\text{gc}},
\end{aligned}
\label{eq:full_loss}
\end{equation}
where $\mathcal{L}_{\text{rgb}}$ and $\mathcal{L}_{\text{d}}$ are the L1 losses between the rendered frame and the GT frame, and between the rendered depth and the GT depth, respectively.
Furthermore, in the main training stage, we compute an additional photometric consistency loss $\mathcal{L}_{\text{d-pc}}$ that utilizes the rendered depth $\hat{\bm{d}}$ of the 3D Gaussians instead of the metric depth~\cite{piccinelli2024unidepth} prior $\bm{d}$ as $\hat{\bm{d}}$ allows the estimated 3D Gaussian geometry to guide the joint optimization of the camera parameter estimation and 3D Gaussian attributes.

In addition to the camera parameter estimation and imagery reconstruction losses, we adopt a binary dice loss~\cite{10.1007/978-3-319-67558-9_28} $\mathcal{L}_\text{M}$ between the pre-computed motion mask $\bm{M}_t$ \cite{yang2023track} and the rendered motion mask $\hat{\bm{M}}_t$ that can be derived from the dynamic 3D Gaussians. The binary dice loss initially proposed in \cite{10.1007/978-3-319-67558-9_28} for highly imbalanced segmentation of medical imagery helps encouraging better separation between our dynamic and static 3D Gaussians as described by
\vspace{-0.1cm}
\begin{equation}
    \mathcal{L}_\text{M} = 1- \frac{2(\textstyle \sum_{\bm{\varphi}_t} \bm{M}_t(\bm{\varphi}_t) \hat{\bm{M}}_t(\bm{\varphi}_t))  + \varepsilon}{(\textstyle \sum_{\bm{\varphi}_t} \bm{M}_t(\bm{\varphi}_t) + \hat{\bm{M}}_t(\bm{\varphi}_t))  + \varepsilon},
    \label{eq:binary_dice_loss}
    \vspace{-0.1cm}
\end{equation}
where $\varepsilon$ is a smooth term to avoid numerical issues. $\hat{\bm{M}}_t(\bm{\varphi}_t)$ is computed by the alpha-blending of the 3D Gaussians overlapping $\bm{\varphi}_t$ (similar to Eq. \ref{eq:alpha_blending}) as
\begin{equation}
    \hat{\bm{M}}_t(\bm{\varphi}_t) = \textstyle \sum_{i\in\mathcal{N}}m_i\alpha_i\prod^{i-1}_{j=1}(1-\alpha_j),
\end{equation}
where $m_i=0$ if the $i^{\text{th}}$ 3D Gaussian is the static 3D Gaussian and $m_i=1$ otherwise.

In conjunction, all terms in $\mathcal{L}^{\text{main}}_{\text{total}}$ guide our SplineGS to effectively and efficiently model dynamic 3D scenes from pure monocular videos, achieving more structural details and better temporal consistency than previous works~\cite{lei2024moscadynamicgaussianfusion,liu2023robust, Wu_2024_CVPR, Li_STG_2024_CVPR, yang2023deformable3dgs, lee2023casual-fvs, huang2023sc, Gao-ICCV-DynNeRF, 23iccv/tian_mononerf, lee2024fully}, without relying on camera parameters obtained from external estimators~\cite{schonberger2016structure}.

\begin{table*}
\begin{center}
\setlength\tabcolsep{6pt} 
\renewcommand{\arraystretch}{1.1}
\scalebox{0.64}{
\begin{tabular}{ c | l | c c c c c c c | c | c}
\toprule
\textbf{PSNR$\uparrow$ / LPIPS$\downarrow$} & \multicolumn{1}{c|}{Method} & Jumping & Skating & Truck & Umbrella & Balloon1 & Balloon2 & Playground & \textbf{Average} & \textbf{FPS}$\uparrow$ \\  
\bottomrule
\hline\noalign{\smallskip}
\multirow{9}{*}{COLMAP} &DynNeRF (ICCV'21)~\cite{Gao-ICCV-DynNeRF} & 24.68 / 0.090 & 32.66 / 0.035 & 28.56 / 0.082 & 23.26 / 0.137 & 22.36 / 0.104 & 27.06 / 0.049 & 24.15 / 0.080 & 26.10 / 0.082 & 0.05 \\
&MonoNeRF (ICCV'23)~\cite{23iccv/tian_mononerf} & 24.26 / 0.091 & 32.06 / 0.044 & 27.56 / 0.115 & 23.62 / 0.180 & 21.89 / 0.129 & 27.36 / 0.052 & 22.61 / 0.130 & 25.62 / 0.106 & 0.05 \\
 &STGS (CVPR'24)~\cite{Li_STG_2024_CVPR}       & 20.82 / 0.187 & 24.80 / 0.109 & 25.01 / 0.103 & 21.88 / 0.195 & 20.36 / 0.196 & 23.12 / 0.124 & 19.23 / 0.151 & 22.17 / 0.152 & \best{900}  \\
 &SCGS (CVPR'24)~\cite{huang2023sc}       & 15.68 / 0.920 & 14.88 / 0.908 & 23.81 / 0.140 & 21.84 / 0.160 & 20.17 / 0.179 & 21.07 / 0.149 & 20.71 / 0.115 & 19.74 / 0.367 & 110  \\
&D3DGS (CVPR'24)~\cite{yang2023deformable3dgs}   & 22.02 / 0.266 & 24.06 / 0.227 & 23.04 / 0.247 & 22.67 / 0.192 & 21.22 / 0.202 & 25.86 / 0.118 & 22.30 / 0.111 & 23.02 / 0.195 & 25  \\
&4DGS (CVPR'24)~\cite{Wu_2024_CVPR} & 22.37 / 0.178 & 26.72 / 0.084 & 25.93 / 0.097 & 22.36 / 0.178 & 21.89 / 0.153 & 24.85 / 0.081 & 21.36 / 0.089 & 23.64 / 0.123 & 95 \\
&RoDynRF (CVPR'23)~\cite{liu2023robust}      & \best{25.66} / \second{0.071} & 28.68 / 0.040 & \best{29.13} / \second{0.063} & 24.26 / \second{0.089} & 22.37 / 0.103 & 26.19 / 0.054 & \second{24.96} / \second{0.048} & 25.89 / \second{0.067} & 0.45 \\
&Casual-FVS (ECCV'24)~\cite{lee2023casual-fvs}   & 23.45 / 0.100 & 29.98 / 0.045 & 25.22 / 0.090 & 23.24 / 0.096 & \second{23.76} / \second{0.079} & 24.15 / 0.081 & 22.19 / 0.074 & 24.57 / 0.081 & 48 \\
&Ex4DGS (NeurIPS'24)~\cite{lee2024fully}   & 18.93 / 0.321 & 21.92 / 0.233 & 19.04 / 0.308 & 19.03 / 0.340 & 14.69 / 0.503 & 16.29 / 0.457 & 14.16 / 0.437 & 17.72 / 0.371 & 84 \\
&MoSca (arXiv)~\cite{lei2024moscadynamicgaussianfusion}       & 25.21 / 0.083 & \second{32.77} / \second{0.033} & 28.22 / 0.090 & \second{24.41} / 0.092 & 23.26 / 0.092 & \second{28.90} / \second{0.042} & 23.05 / 0.060 & 26.55 / 0.070 & N/A \\
\midrule
\multirow{3}{*}{COLMAP-Free} & RoDynRF (CVPR'23)~\cite{liu2023robust}      & 24.27 / 0.100 & 28.71 / 0.046 & \second{28.85} / 0.066 & 23.25 / 0.104 & 21.81 / 0.122 & 25.58 / 0.064 & \best{25.20} / 0.052 & 25.38 / 0.079 & 0.45 \\
&MoSca (arXiv)~\cite{lei2024moscadynamicgaussianfusion}       & 25.43 / 0.080 & 32.62 / \second{0.033} & 28.29 / 0.086 & 24.40 / 0.091 & 23.27 / 0.091 & \best{29.01} / \second{0.042} & 23.23 / 0.058 & \second{26.61} / 0.069 & N/A \\
&\textbf{SplineGS (Ours)} & \second{25.50} / \best{0.068} & \best{33.72} / \best{0.031} & 28.66 / \best{0.056} & \best{25.61} / \best{0.071} & \best{24.43} / \best{0.068} & 28.37 / \best{0.032} & 24.19 / \best{0.047} & \best{27.21} / \best{0.053} & \second{400}  \\
\bottomrule
\end{tabular}
}
\vspace{-0.2cm}
\caption{\textbf{Novel view synthesis evaluation on the NVIDIA dataset.} \best{Red} and \second{Blue} denote the best and second-best performances, respectively. `N/A' denotes that the rendering speed for MoSca~\cite{lei2024moscadynamicgaussianfusion} is unavailable, as the authors have not provided official code. For Casual-FVS~\cite{lee2023casual-fvs}, we directly use the results from their paper, as official code is also unavailable.}
\label{table:nvidia_quantitative}
\vspace{-0.4cm}
\end{center}
\end{table*}
\begin{figure*}[t]
    \centering
    \includegraphics[width=\linewidth,keepaspectratio]{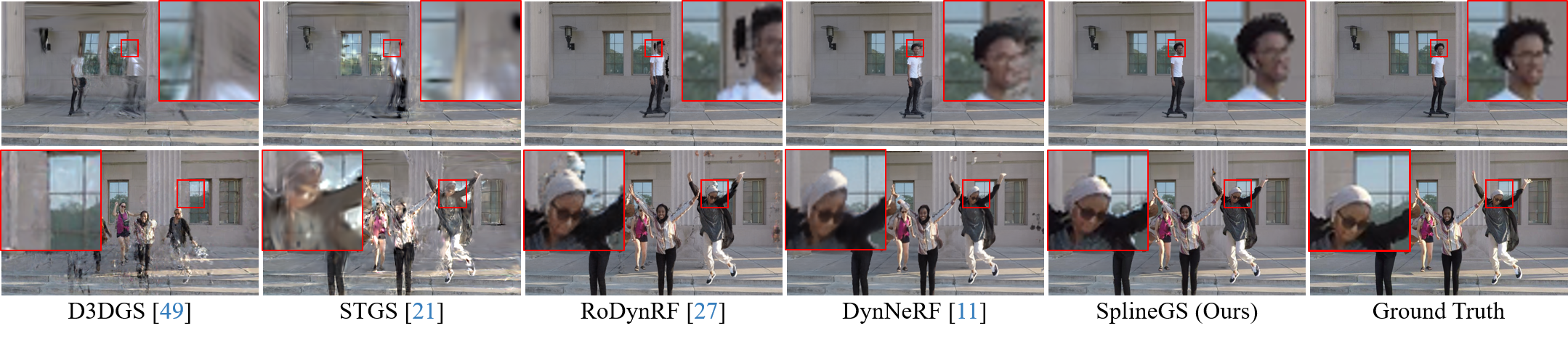}
    \vspace{-0.8cm}
    \caption{\textbf{Visual comparisons for novel view synthesis on the NVIDIA dataset.}}
    \label{fig:qualitative_nvidia}
    \vspace{-0.5cm}
\end{figure*}

\section{Experiments}
\label{sec:experiments}

\noindent \textbf{Implementation Details.}
To develop our method, we build on top of the widely used open-source 3DGS codebase \cite{kerbl20233d}. Our SplineGS architecture is trained over 1K iterations in the warm-up stage and 20K iterations in the main training stage. We optimize the number of control points with the proposed MACP every 100 iterations. For depth and 2D tracking estimation, we employ the pre-trained models from UniDepth~\cite{piccinelli2024unidepth} and CoTracker~\cite{karaev2023cotracker}, respectively. The learnable camera extrinsics $[\hat{\bm{R}}_t|\hat{\bm{T}}_t]$ are initialized by $[\textbf{I} | \bm{0}]$, while the initial learnable focal length value $\hat{f}$ is set to 500.

\noindent \textbf{Datasets.} We evaluate both the quantitative and qualitative performance of novel view and time synthesis on the widely used NVIDIA dataset \cite{yoon2020dynamic} which features challenging monocular videos. Additionally, we assess novel view synthesis performance on in-the-wild monocular videos from the DAVIS dataset \cite{ponttuset20182017davischallengevideo} which contains an average of 70 frames per video sequence.




\subsection{Comparison with State-of-the-Art Methods}
\label{sec:quanti}

\noindent \textbf{Novel View Synthesis.} Table \ref{table:nvidia_quantitative} presents a quantitative comparison of NVS between our SplineGS and existing COLMAP-based \cite{Gao-ICCV-DynNeRF, 23iccv/tian_mononerf, Li_STG_2024_CVPR, huang2023sc, yang2023deformable3dgs, Wu_2024_CVPR, lee2023casual-fvs, lee2024fully} and COLMAP-free \cite{liu2023robust, lei2024moscadynamicgaussianfusion} methods on the NVIDIA dataset \cite{yoon2020dynamic}. For this comparison, we follow the evaluation configuration in \cite{liu2023robust}. The results demonstrate that our SplineGS significantly outperforms SOTA methods in both the PSNR and LPIPS \cite{zhang2018perceptual} metrics. Notably, SplineGS achieves \textbf{890}$\times$ and \textbf{8,000}$\times$ faster rendering speed compared to RoDynRF \cite{liu2023robust} and DynNeRF \cite{Gao-ICCV-DynNeRF}, respectively. Ex4DGS \cite{lee2024fully} and STGS \cite{Li_STG_2024_CVPR}, which are designed for multi-view settings, face challenges with inconsistent geometry alignment over time when trained on monocular videos. Furthermore, the LPIPS score of our SplineGS is consistently superior to all the methods across all scenes. 

\begin{figure}
    \centering
    \includegraphics[width=\linewidth,keepaspectratio]{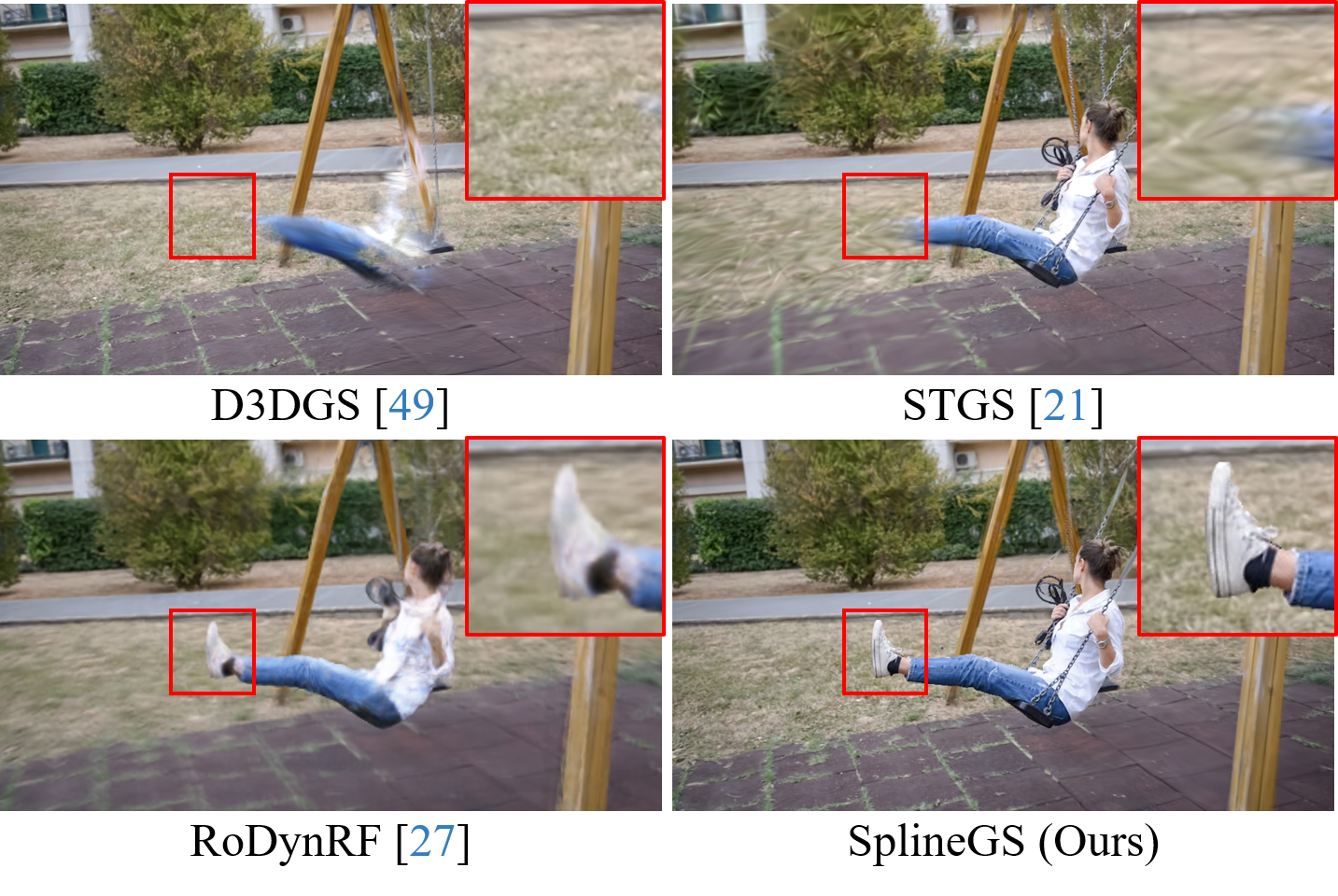}
    \vspace{-0.8cm}
    \caption{\textbf{Visual comparisons for novel view synthesis on the DAVIS dataset.}}
    \label{fig:qualitative_davis}
    \vspace{-0.8cm}
\end{figure}

Fig. \ref{fig:qualitative_nvidia} shows the qualitative comparison between our SplineGS with the existing methods in \cite{yang2023deformable3dgs, Li_STG_2024_CVPR, Gao-ICCV-DynNeRF, liu2023robust}. As highlighted by the red boxes, our method yields not only higher rendering quality but also dynamic objects more aligned and closer to the ground truth. In Fig. \ref{fig:qualitative_davis}, we show our superior NVS results for in-the-wild monocular videos from the DAVIS dataset \cite{ponttuset20182017davischallengevideo} compared to the existing methods in \cite{yang2023deformable3dgs, Li_STG_2024_CVPR, liu2023robust}. Compared to~\cite{liu2023robust} that is also COLMAP-free, our SplineGS yields considerably more detailed novel views (red boxes) as shown in Fig.~\ref{fig:qualitative_davis}. For the other methods~\cite{Li_STG_2024_CVPR, yang2023deformable3dgs}, we observe that COLMAP~\cite{schonberger2016structure} fails to recover camera parameters and initial point clouds on the DAVIS dataset \cite{ponttuset20182017davischallengevideo}, as also claimed in~\cite{liu2023robust}. On the other hand, our COLMAP-free SplineGS reconstructs accurate camera parameters that are the ones actually used to train \cite{yang2023deformable3dgs, Li_STG_2024_CVPR}, which are shown in Fig.~\ref{fig:qualitative_davis} for comparison. More results on DAVIS \cite{ponttuset20182017davischallengevideo} are provided in the \textit{Suppl}.

\begin{figure}[t]
    \centering
    \includegraphics[width=\linewidth,keepaspectratio]{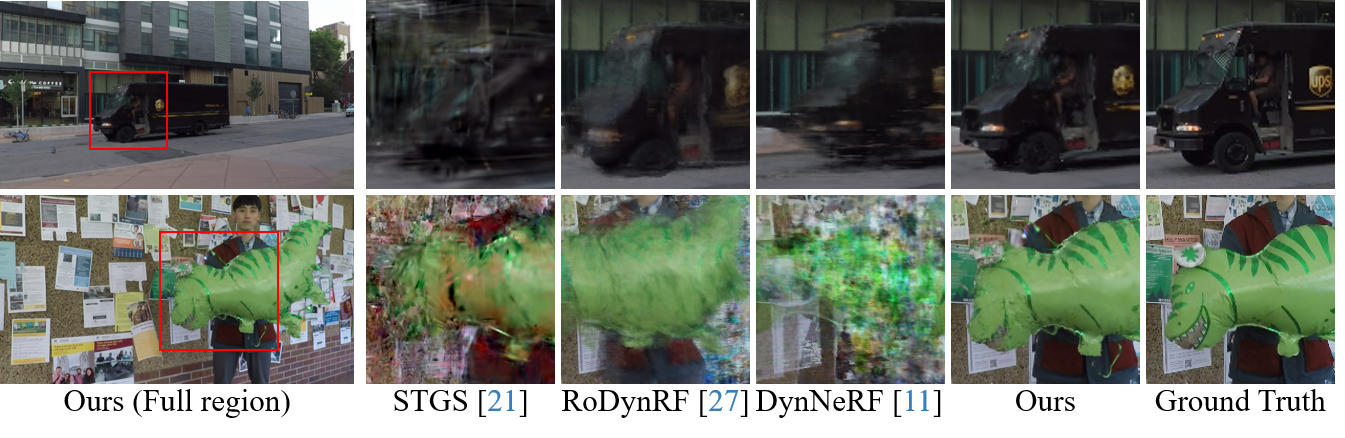}
    \vspace{-0.7cm}
    \caption{\textbf{Visual comparisons for novel view and time synthesis on the NVIDIA dataset.}}
    \label{fig:qualitative_t_interp}
    \vspace{-0.5cm}
\end{figure}

\noindent \textbf{Novel View and Time Synthesis.} To evaluate the capability of SplineGS to model continuous trajectories of moving objects in a scene, we compare the novel view and time synthesis results of SplineGS with those of NeRF-based \cite{Gao-ICCV-DynNeRF, liu2023robust} and 3DGS-based \cite{Li_STG_2024_CVPR, Wu_2024_CVPR, yang2023deformable3dgs} methods.
For this evaluation, we follow the dataset sampling strategy in \cite{li2021neural}, which samples 24 timestamps from the NVIDIA dataset~\cite{yoon2020dynamic}. In addition, to simulate a larger motion, we exclude frames with odd time indices in the training sets.
To ensure all test timestamps are not seen during training, and thus, to create a more \textit{challenging} novel view and time synthesis validation, we exclude frames with even time indices in the test sets. Table~\ref{table:nvidialow_quantitative} and Fig. \ref{fig:qualitative_t_interp} show the quantitative and qualitative comparisons for this challenging experiment. 
We observe that the NeRF-based methods, RoDynRF~\cite{liu2023robust} and DynNeRF~\cite{Gao-ICCV-DynNeRF}, generate inconsistent artifacts and blurriness for unseen times. Furthermore, the 3DGS-based methods \cite{yang2023deformable3dgs, Wu_2024_CVPR, Li_STG_2024_CVPR} yield even more significant degradation when predicting unseen time indices. In contrast, SplineGS, built upon 3DGS~\cite{kerbl20233d} but equipped with our novel spline-based deformation, provides SOTA novel view rendering for unseen intermediate time. Thanks to our MAS, SplineGS naturally and precisely captures the continuous trajectories of moving objects over time, enhancing the temporal consistency of rendered scenes that can be checked in tOF scores \cite{chu2020learning} of Table~\ref{table:nvidialow_quantitative}.

\begin{figure}[h]
    \centering
    \includegraphics[width=\linewidth,keepaspectratio]{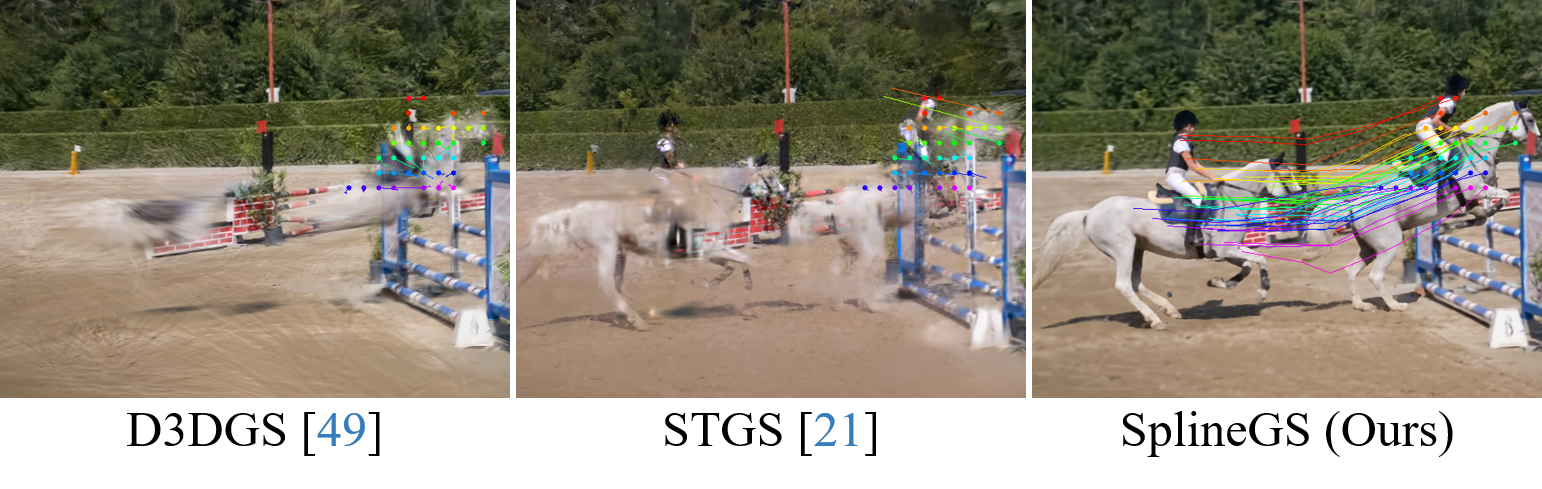}
    \vspace{-0.8cm}
    \caption{\textbf{Visual comparisons for motion tracking.} We visualize 2D pixel tracks to analyze motions of dynamic 3D Gaussians.}
    \label{fig:tracking}
    \vspace{-0.5cm}
\end{figure}

To further analyze the SplineGS's ability to model continuous trajectories of dynamic 3D Gaussians, we visualize the projected 2D motion tracking of dynamic objects in pixel space, comparing it with D3DGS~\cite{yang2023deformable3dgs} and STGS~\cite{Li_STG_2024_CVPR}, as shown in Fig.~\ref{fig:tracking}. We observe that D3DGS~\cite{yang2023deformable3dgs} and STGS~\cite{Li_STG_2024_CVPR} cannot provide reliable motion tracking for moving objects, underscoring their limitations in modeling continuous trajectories of dynamic 3D Gaussians. In contrast, SplineGS provides accurate motion tracking, demonstrating the effectiveness of our MAS for deforming dynamic 3D Gaussians. More results are provided in \textit{Suppl}.

\begin{table}
\begin{center}
\scalebox{0.7}{
\begin{tabular}{ c | l | c c c }
\toprule
& \multicolumn{1}{c|}{Method} & PSNR$\uparrow$ & LPIPS$\downarrow$ & tOF$\downarrow$ \\  
\bottomrule
\hline\noalign{\smallskip}
\multirow{4}{*}{COLMAP}&DynNeRF (ICCV'21)~\cite{Gao-ICCV-DynNeRF}     & \second{23.36} & \second{0.219} & \second{0.921} \\
 &4DGS (CVPR'24)~\cite{Wu_2024_CVPR}     & 17.07 & 0.459 & 6.314 \\
&D3DGS (CVPR'24)~\cite{yang2023deformable3dgs}  & 19.63 &  0.343 & 3.225\\
&STGS (CVPR'24)~\cite{Li_STG_2024_CVPR}  & 15.72 &  0.474 & 2.105 \\
\midrule
\multirow{2}{*}{COLMAP-Free} & RoDynRF (CVPR'23)~\cite{liu2023robust}       & 21.58 & 0.221  & 2.138 \\
&\textbf{SplineGS (Ours)} & \best{25.92} & \best{0.098} & \best{0.703} \\
\bottomrule
\end{tabular}
}
\vspace{-0.2cm}
\caption{\textbf{Novel view and time synthesis evaluation on the NVIDIA dataset.}}
\label{table:nvidialow_quantitative}
\vspace{-0.4cm}
\end{center}
\end{table}
\vspace{-0.3cm}

\subsection{Ablation Study}
\label{sec:ablation}
\noindent \textbf{Motion-Adaptive Spline (MAS).} To demonstrate the effectiveness of MAS, we replace the MAS model with various deformation models, including an MLP, a grid-based model, polynomial functions of third degree (denoted as `Poly ($3^{\text{rd}}$)') and tenth degree (denoted as `Poly ($10^{\text{th}}$)'), and a Bézier curve~\cite{farin2001curves}. For the MLP, grid-based model, and polynomial functions, we apply to them the structures similar to those in prior works, including D3DGS \cite{yang2023deformable3dgs}, 4DGS \cite{Wu_2024_CVPR}, and STGS \cite{Li_STG_2024_CVPR}, respectively. Additionally, we implement the Bézier curve~\cite{farin2001curves}, a commonly used method for curve modeling in computer graphics.
Table \ref{table:ablation_study}-(a) presents quantitative comparisons of each 3D Gaussian trajectory model, focusing on rendering quality (PSNR, LPIPS) and deformation latency per Gaussian, denoted as $g_{\text{def}}$. This latency reflects the computational time required to estimate the deformation of a single dynamic 3D Gaussian. As shown in Table \ref{table:ablation_study}-(a), our MAS model achieves superior rendering quality compared to all other deformation models. Consistent with analyses in previous works \cite{Wu_2024_CVPR, Li_STG_2024_CVPR, yang2023deformable3dgs}, we observe that the MLP and grid-based architectures require substantial computational costs for rendering.
Among these methods, `Poly ($3^{\text{rd}}$)', as implemented in \cite{Li_STG_2024_CVPR}, demonstrates the best latency. However, fixed-degree polynomial functions have limited flexibility across varying motion complexities which adversely impacts rendering performance. To explore this further, we experiment with `Poly ($10^{\text{th}}$)' to assess changes in modeling capability. This adjustment, however, leads to noisier optimization and reduced efficiency, as variables under high exponents in `Poly ($10^{\text{th}}$)' lead to numerical instability. The Bézier curve~\cite{farin2001curves} offers the second-best rendering quality, but its latency remains higher than our MAS due to its recursive nature of computation.

\begin{figure}[t]
    \centering
    \includegraphics[width=\linewidth,keepaspectratio]{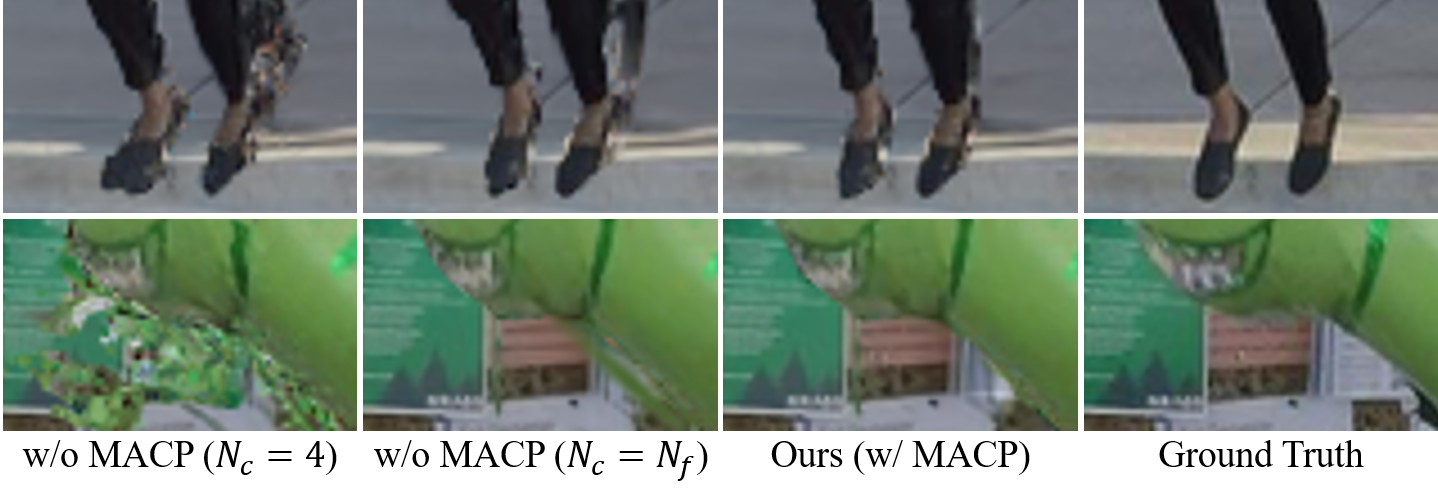}
    \vspace{-0.8cm}
    \caption{\textbf{Visual comparisons for MACP ablation study.}}
    \label{fig:macp_n_c}
    \vspace{-0.5cm}
\end{figure}

\begin{figure}[t]
    \centering
    \includegraphics[width=\linewidth,keepaspectratio]{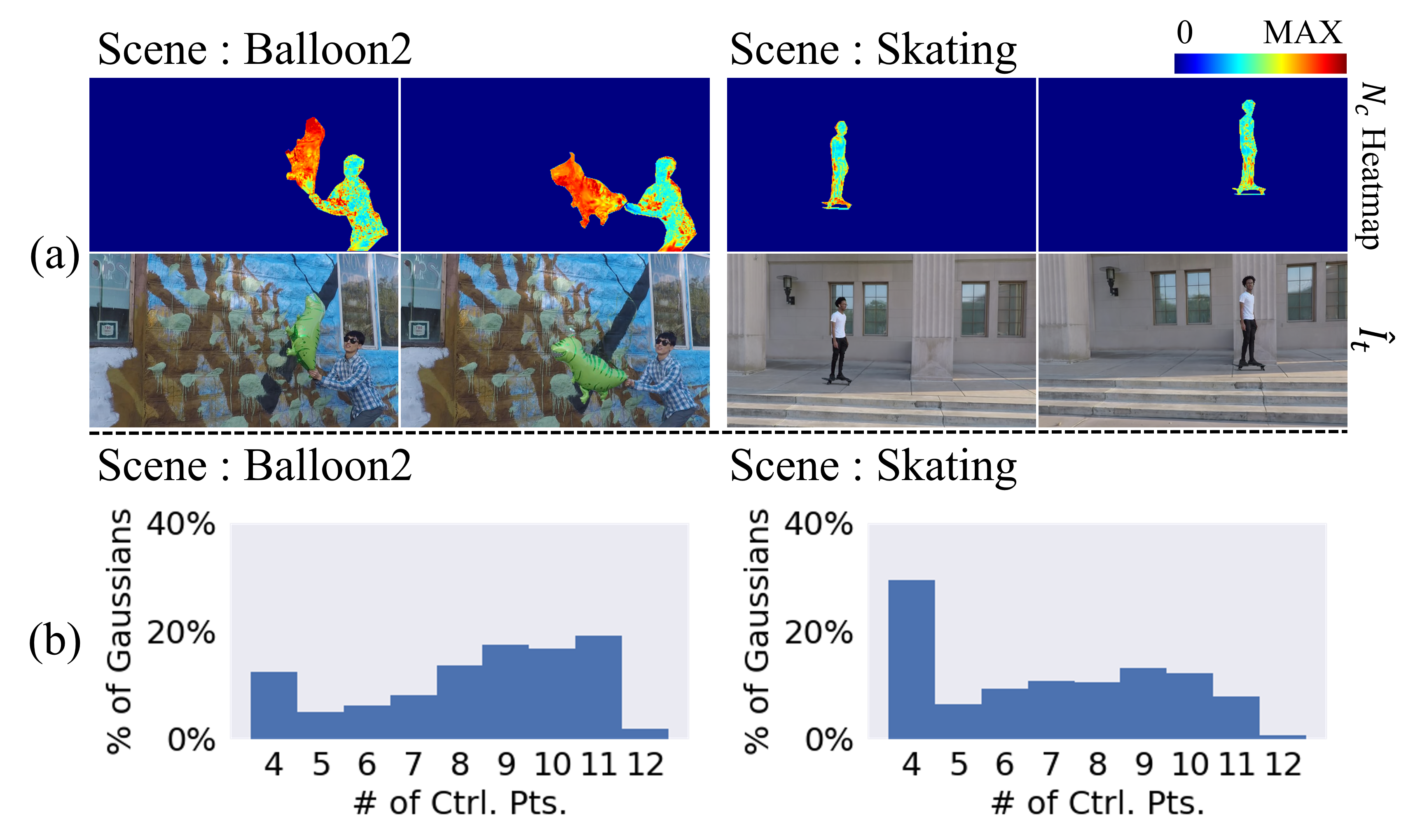}
    \vspace{-0.7cm}
    \caption{\textbf{Analysis of MACP's Efficacy.} (a) $N_c$ Heatmaps as the averaged $N_c$ values of dynamic 3D Gaussians and their corresponding rendered frames $\hat{I}_t$ for `Balloon2' and `Skating' scenes. (b) Histograms of the number of control points ($N_c$) in percentages (\%) of dynamic 3D Gaussians in two scenes.}
    \label{fig:macp}
    \vspace{-0.5cm}
\end{figure}

\noindent \textbf{Motion-Adaptive Control Points Pruning (MACP).}
To assess the effectiveness of our MACP technique for MAS, we compare our full model with MACP against other versions of our model with two fixed numbers of control points $N_c = 4$ and $N_c = N_f$. As shown in Table \ref{table:ablation_study}-(c) and Fig. \ref{fig:macp_n_c}, our SplineGS with MACP achieves a good trade-off between rendering quality and $g_{\text{def}}$ compared to the ablated models with fixed $N_c$. Using $N_c=4$ for every dynamic 3D Gaussian limits the motion modeling capacity of MAS, resulting in significantly lower metrics and visible artifacts in the dynamic regions. Moreover, an excessive $N_c = N_f$ decreases the rendering speed of our MAS module and still falls short of the quality achieved by our full model with MACP, potentially due to motion overfitting.
Fig.~\ref{fig:macp} shows the distribution of $N_c$ values after optimization with MACP across scenes of varying motion complexities. In Fig.~\ref{fig:macp}-(a), we visualize `$N_c$ Heatmap' that contains the pixel-wise averaged $N_c$ values of the dynamic 3D Gaussians needed to render the 2D pixels (red higher, blue lower number of control points). As shown, the simpler motions in these scenes, such as those of the human bodies, can be modeled by the dynamic 3D Gaussians with smaller averages of $N_c$ values, whereas the objects with complex and extensive motions (e.g. balloon) require higher averaged $N_c$ values. Fig.~\ref{fig:macp}-(b) presents the corresponding histogram of $N_c$ values for the scenes' dynamic 3D Gaussians. For sequences with simple motion, such as ‘Skating’, the trajectories of most dynamic 3D Gaussians can be represented using a minimal $N_c$, thanks to our MACP. While ‘Balloon2’ has more evenly distributed $N_c$ due to more complex and diverse motion.

\begin{table}[t]
\begin{minipage}[t]{0.5\linewidth}
\centering
\captionsetup{justification=centering}
\caption*{(a) Motion-Adaptive Spline}
\footnotesize
\vspace{-0.3cm}
\scalebox{0.7}{
\begin{tabular}{ l| c c c }
\bottomrule
\hline\noalign{\smallskip}
   & PSNR$\uparrow$ & LPIPS$\downarrow$ & $g_{\text{def}}$ (ns)$\downarrow$\\  
\bottomrule
\hline\noalign{\smallskip}
  MLP &  23.51  & 0.125 & 149.41  \\
  Grid    &  25.48 & 0.090 & 98.89 \\
  Poly ($3^{\text{rd}}$)     & 25.14 & 0.111 &   \best{1.80}    \\
  Poly ($10^{\text{th}}$) & 24.38 & 0.120 &  7.71 \\
  Bézier & \second{27.19} & \second{0.060} &  8.78 \\
\midrule
 Ours  & \best{27.21} & \best{0.053} & \second{5.63} \\
\bottomrule
\hline\noalign{\smallskip}
\end{tabular}
}
\end{minipage}%
\begin{minipage}[t]{0.5\linewidth}
\centering
\captionsetup{justification=centering}

\caption*{(b) Loss function}
\footnotesize
\vspace{-0.3cm}
\scalebox{0.79}{
\begin{tabular}{ l| c c }
\bottomrule
\hline\noalign{\smallskip}
   & PSNR$\uparrow$ & LPIPS$\downarrow$ \\  
\bottomrule
\hline\noalign{\smallskip}
w/o $\mathcal{L}_\text{pc}$      & 17.49 & 0.853 \\
w/o $\mathcal{L}_\text{gc}$      & 26.33 & 0.067 \\
w/o $\mathcal{L}_\text{d-pc}$    & 26.18 & \second{0.066} \\
w/o $\mathcal{L}_\text{M}$       & \second{26.34} & 0.088 \\
\midrule
Ours & \best{27.21} & \best{0.053} \\
\bottomrule
\hline\noalign{\smallskip}
\end{tabular}}
\end{minipage}

\begin{minipage}[t]{\linewidth}
\centering
\captionsetup{justification=centering}
\caption*{(c) Motion-Adaptive Control points Pruning}
\footnotesize
\vspace{-0.3cm}
\scalebox{0.79}{
\begin{tabular}{ l| c c c }
\bottomrule
\hline\noalign{\smallskip}
   & PSNR$\uparrow$ & LPIPS$\downarrow$ & $g_{\text{def}}$ (ns)$\downarrow$\\  
\bottomrule
\hline\noalign{\smallskip}
  w/o MACP ($N_c=4$)       & 26.62 & 0.065 & \best{5.34}  \\
  w/o MACP ($N_c = N_f$)       & \second{27.08} & \second{0.054} &  6.11 \\
\midrule
 Ours  & \best{27.21} & \best{0.053} & \second{5.63} \\
\bottomrule
\hline\noalign{\smallskip}
\end{tabular}
}
\end{minipage}
\vspace{-0.3cm}
\caption{\textbf{Ablation studies.} We ablate our framework and report the average results on the NVIDIA dataset with the same setting as Novel View Synthesis experiment in Sec. \ref{sec:quanti}.}
\vspace{-0.2cm}
\label{table:ablation_study}
\end{table}

\noindent \textbf{Loss Functions.} Table \ref{table:ablation_study}-(b) shows the effectiveness of each loss for our overall SplineGS architecture. As noted, no consistent camera parameters can be learned without $\mathcal{L}_\text{pc}$, drastically impacting the rendering quality of the dynamic 3D Gaussians. Also, our $\mathcal{L}_\text{gc}$, $\mathcal{L}_\text{d-pc}$ and $\mathcal{L}_\text{M}$ can considerably impact the overall rendering quality.
\section{Conclusion}
We present SplineGS, a COLMAP-free dynamic 3DGS framework designed for novel spatio-temporal view synthesis from monocular videos. Leveraging our innovative Motion-Adaptive Spline (MAS) for dynamic motion modeling, SplineGS efficiently renders high-quality novel views from complex in-the-wild videos. The effectiveness of our approach is validated through extensive quantitative and qualitative comparisons, significantly outperforming the existing SOTA methods with very fast rendering speed.
{
    \small
    \bibliographystyle{ieeenat_fullname}
    \bibliography{main}

\begin{thebibliography}{51}
\providecommand{\natexlab}[1]{#1}
\providecommand{\url}[1]{\texttt{#1}}
\expandafter\ifx\csname urlstyle\endcsname\relax
  \providecommand{\doi}[1]{doi: #1}\else
  \providecommand{\doi}{doi: \begingroup \urlstyle{rm}\Url}\fi

\bibitem[fan(2022)]{fang2022fast}
Fast dynamic radiance fields with time-aware neural voxels.
\newblock In \emph{SIGGRAPH Asia 2022 Conference Papers}, 2022.

\bibitem[Ahlberg et~al.(2016)Ahlberg, Nilson, and Walsh]{ahlberg2016theory}
J~Harold Ahlberg, Edwin~Norman Nilson, and Joseph~Leonard Walsh.
\newblock \emph{The Theory of Splines and Their Applications: Mathematics in Science and Engineering: A Series of Monographs and Textbooks, Vol. 38}.
\newblock Elsevier, 2016.

\bibitem[Athar et~al.(2022)Athar, Xu, Sunkavalli, Shechtman, and Shu]{athar2022rignerf}
ShahRukh Athar, Zexiang Xu, Kalyan Sunkavalli, Eli Shechtman, and Zhixin Shu.
\newblock Rignerf: Fully controllable neural 3d portraits.
\newblock In \emph{CVPR}, 2022.

\bibitem[Attal et~al.(2023)Attal, Huang, Richardt, Zollhoefer, Kopf, O’Toole, and Kim]{attal2023hyperreel}
Benjamin Attal, Jia-Bin Huang, Christian Richardt, Michael Zollhoefer, Johannes Kopf, Matthew O’Toole, and Changil Kim.
\newblock Hyperreel: High-fidelity 6-dof video with ray-conditioned sampling.
\newblock In \emph{CVPR}, 2023.

\bibitem[Cao and Johnson(2023)]{cao2023hexplane}
Ang Cao and Justin Johnson.
\newblock Hexplane: A fast representation for dynamic scenes.
\newblock In \emph{CVPR}, 2023.

\bibitem[Chu et~al.(2020)Chu, Xie, Mayer, Leal-Taix{\'e}, and Thuerey]{chu2020learning}
Mengyu Chu, You Xie, Jonas Mayer, Laura Leal-Taix{\'e}, and Nils Thuerey.
\newblock Learning temporal coherence via self-supervision for gan-based video generation.
\newblock \emph{ACM Transactions on Graphics (TOG)}, 2020.

\bibitem[De~Boor(1978)]{de1978practical}
C De~Boor.
\newblock A practical guide to splines.
\newblock \emph{Springer-Verlag google schola}, 1978.

\bibitem[Farin(2001)]{farin2001curves}
Gerald Farin.
\newblock \emph{Curves and surfaces for CAGD: a practical guide}.
\newblock Elsevier, 2001.

\bibitem[Fridovich-Keil et~al.(2023)Fridovich-Keil, Meanti, Warburg, Recht, and Kanazawa]{fridovich2023k}
Sara Fridovich-Keil, Giacomo Meanti, Frederik~Rahb{\ae}k Warburg, Benjamin Recht, and Angjoo Kanazawa.
\newblock K-planes: Explicit radiance fields in space, time, and appearance.
\newblock In \emph{CVPR}, 2023.

\bibitem[Fu et~al.(2024)Fu, Liu, Kulkarni, Kautz, Efros, and Wang]{Fu_2024_CVPR}
Yang Fu, Sifei Liu, Amey Kulkarni, Jan Kautz, Alexei~A. Efros, and Xiaolong Wang.
\newblock Colmap-free 3d gaussian splatting.
\newblock In \emph{CVPR}, 2024.

\bibitem[Gao et~al.(2021)Gao, Saraf, Kopf, and Huang]{Gao-ICCV-DynNeRF}
Chen Gao, Ayush Saraf, Johannes Kopf, and Jia-Bin Huang.
\newblock Dynamic view synthesis from dynamic monocular video.
\newblock In \emph{ICCV}, 2021.

\bibitem[Horn and Johnson(2012)]{horn2012matrix}
Roger~A Horn and Charles~R Johnson.
\newblock \emph{Matrix analysis}.
\newblock Cambridge university press, 2012.

\bibitem[Huang et~al.(2024)Huang, Sun, Yang, Lyu, Cao, and Qi]{huang2023sc}
Yi-Hua Huang, Yang-Tian Sun, Ziyi Yang, Xiaoyang Lyu, Yan-Pei Cao, and Xiaojuan Qi.
\newblock Sc-gs: Sparse-controlled gaussian splatting for editable dynamic scenes.
\newblock \emph{CVPR}, 2024.

\bibitem[Jiang et~al.(2022)Jiang, Yi, Samei, Tuzel, and Ranjan]{jiang2022neuman}
Wei Jiang, Kwang~Moo Yi, Golnoosh Samei, Oncel Tuzel, and Anurag Ranjan.
\newblock Neuman: Neural human radiance field from a single video.
\newblock In \emph{ECCV}, 2022.

\bibitem[Jiang et~al.(2023)Jiang, Hedman, Mildenhall, Xu, Barron, Wang, and Xue]{jiang2023alignerf}
Yifan Jiang, Peter Hedman, Ben Mildenhall, Dejia Xu, Jonathan~T Barron, Zhangyang Wang, and Tianfan Xue.
\newblock Alignerf: High-fidelity neural radiance fields via alignment-aware training.
\newblock In \emph{CVPR}, 2023.

\bibitem[Karaev et~al.(2023)Karaev, Rocco, Graham, Neverova, Vedaldi, and Rupprecht]{karaev2023cotracker}
Nikita Karaev, Ignacio Rocco, Benjamin Graham, Natalia Neverova, Andrea Vedaldi, and Christian Rupprecht.
\newblock Cotracker: It is better to track together.
\newblock \emph{arXiv}, 2023.

\bibitem[Kerbl et~al.(2023)Kerbl, Kopanas, Leimk{\"u}hler, and Drettakis]{kerbl20233d}
Bernhard Kerbl, Georgios Kopanas, Thomas Leimk{\"u}hler, and George Drettakis.
\newblock 3d gaussian splatting for real-time radiance field rendering.
\newblock \emph{ACM Trans. Graph.}, 2023.

\bibitem[Lee et~al.(2024{\natexlab{a}})Lee, Won, Jung, Bae, and Jeon]{lee2024fully}
Junoh Lee, Chang-Yeon Won, Hyunjun Jung, Inhwan Bae, and Hae-Gon Jeon.
\newblock Fully explicit dynamic gaussian splatting.
\newblock In \emph{NeurIPS}, 2024{\natexlab{a}}.

\bibitem[Lee et~al.(2024{\natexlab{b}})Lee, Zhang, Blackburn-Matzen, Niklaus, Zhang, Huang, and Liu]{lee2023casual-fvs}
Yao-Chih Lee, Zhoutong Zhang, Kevin Blackburn-Matzen, Simon Niklaus, Jianming Zhang, Jia-Bin Huang, and Feng Liu.
\newblock Fast view synthesis of casual videos with soup-of-planes.
\newblock In \emph{ECCV}, 2024{\natexlab{b}}.

\bibitem[Lei et~al.(2024)Lei, Weng, Harley, Guibas, and Daniilidis]{lei2024moscadynamicgaussianfusion}
Jiahui Lei, Yijia Weng, Adam Harley, Leonidas Guibas, and Kostas Daniilidis.
\newblock Mosca: Dynamic gaussian fusion from casual videos via 4d motion scaffolds.
\newblock \emph{arXiv}, 2024.

\bibitem[Li et~al.()Li, Chen, Li, and Xu]{Li_STG_2024_CVPR}
Zhan Li, Zhang Chen, Zhong Li, and Yi Xu.
\newblock Spacetime gaussian feature splatting for real-time dynamic view synthesis.
\newblock In \emph{CVPR}.

\bibitem[Li et~al.(2021)Li, Niklaus, Snavely, and Wang]{li2021neural}
Zhengqi Li, Simon Niklaus, Noah Snavely, and Oliver Wang.
\newblock Neural scene flow fields for space-time view synthesis of dynamic scenes.
\newblock In \emph{CVPR}, 2021.

\bibitem[Li et~al.(2023)Li, Wang, Cole, Tucker, and Snavely]{li2023dynibar}
Zhengqi Li, Qianqian Wang, Forrester Cole, Richard Tucker, and Noah Snavely.
\newblock Dynibar: Neural dynamic image-based rendering.
\newblock In \emph{CVPR}, 2023.

\bibitem[Liang et~al.(2023)Liang, Khan, Li, Nguyen-Phuoc, Lanman, Tompkin, and Xiao]{liang2023gaufre}
Yiqing Liang, Numair Khan, Zhengqin Li, Thu Nguyen-Phuoc, Douglas Lanman, James Tompkin, and Lei Xiao.
\newblock Gaufre: Gaussian deformation fields for real-time dynamic novel view synthesis.
\newblock \emph{arXiv}, 2023.

\bibitem[Lin et~al.(2021)Lin, Ma, Torralba, and Lucey]{lin2021barf}
Chen-Hsuan Lin, Wei-Chiu Ma, Antonio Torralba, and Simon Lucey.
\newblock Barf: Bundle-adjusting neural radiance fields.
\newblock In \emph{ICCV}, 2021.

\bibitem[Lindenberger et~al.(2021)Lindenberger, Sarlin, Larsson, and Pollefeys]{lindenberger2021pixel}
Philipp Lindenberger, Paul-Edouard Sarlin, Viktor Larsson, and Marc Pollefeys.
\newblock Pixel-perfect structure-from-motion with featuremetric refinement.
\newblock In \emph{ICCV}, 2021.

\bibitem[Liu et~al.(2023)Liu, Gao, Meuleman, Tseng, Saraf, Kim, Chuang, Kopf, and Huang]{liu2023robust}
Yu-Lun Liu, Chen Gao, Andreas Meuleman, Hung-Yu Tseng, Ayush Saraf, Changil Kim, Yung-Yu Chuang, Johannes Kopf, and Jia-Bin Huang.
\newblock Robust dynamic radiance fields.
\newblock In \emph{CVPR}, 2023.

\bibitem[Luiten et~al.(2024)Luiten, Kopanas, Leibe, and Ramanan]{luiten2023dynamic}
Jonathon Luiten, Georgios Kopanas, Bastian Leibe, and Deva Ramanan.
\newblock Dynamic 3d gaussians: Tracking by persistent dynamic view synthesis.
\newblock In \emph{3DV}, 2024.

\bibitem[Meng et~al.(2021)Meng, Chen, Luo, Wu, Su, Xu, He, and Yu]{meng2021gnerf}
Quan Meng, Anpei Chen, Haimin Luo, Minye Wu, Hao Su, Lan Xu, Xuming He, and Jingyi Yu.
\newblock Gnerf: Gan-based neural radiance field without posed camera.
\newblock In \emph{ICCV}, 2021.

\bibitem[Mildenhall et~al.(2020)Mildenhall, Srinivasan, Tancik, Barron, Ramamoorthi, and Ng]{mildenhall2020nerf}
Ben Mildenhall, Pratul~P. Srinivasan, Matthew Tancik, Jonathan~T. Barron, Ravi Ramamoorthi, and Ren Ng.
\newblock Nerf: Representing scenes as neural radiance fields for view synthesis.
\newblock In \emph{ECCV}, 2020.

\bibitem[Park et~al.(2021{\natexlab{a}})Park, Sinha, Barron, Bouaziz, Goldman, Seitz, and Martin-Brualla]{park2021nerfies}
Keunhong Park, Utkarsh Sinha, Jonathan~T Barron, Sofien Bouaziz, Dan~B Goldman, Steven~M Seitz, and Ricardo Martin-Brualla.
\newblock Nerfies: Deformable neural radiance fields.
\newblock In \emph{ICCV}, 2021{\natexlab{a}}.

\bibitem[Park et~al.(2021{\natexlab{b}})Park, Sinha, Hedman, Barron, Bouaziz, Goldman, Martin-Brualla, and Seitz]{park2021hypernerf}
Keunhong Park, Utkarsh Sinha, Peter Hedman, Jonathan~T. Barron, Sofien Bouaziz, Dan~B Goldman, Ricardo Martin-Brualla, and Steven~M. Seitz.
\newblock Hypernerf: A higher-dimensional representation for topologically varying neural radiance fields.
\newblock \emph{ACM Trans. Graph.}, 2021{\natexlab{b}}.

\bibitem[Park et~al.(2023)Park, Henzler, Mildenhall, Barron, and Martin-Brualla]{camp}
Keunhong Park, Philipp Henzler, Ben Mildenhall, Jonathan~T. Barron, and Ricardo Martin-Brualla.
\newblock Camp: Camera preconditioning for neural radiance fields.
\newblock \emph{ACM Trans. Graph.}, 2023.

\bibitem[Piccinelli et~al.(2024)Piccinelli, Yang, Sakaridis, Segu, Li, Van~Gool, and Yu]{piccinelli2024unidepth}
Luigi Piccinelli, Yung-Hsu Yang, Christos Sakaridis, Mattia Segu, Siyuan Li, Luc Van~Gool, and Fisher Yu.
\newblock Unidepth: Universal monocular metric depth estimation.
\newblock In \emph{CVPR}, 2024.

\bibitem[Pont-Tuset et~al.(2018)Pont-Tuset, Perazzi, Caelles, Arbeláez, Sorkine-Hornung, and Gool]{ponttuset20182017davischallengevideo}
Jordi Pont-Tuset, Federico Perazzi, Sergi Caelles, Pablo Arbeláez, Alex Sorkine-Hornung, and Luc~Van Gool.
\newblock The 2017 davis challenge on video object segmentation.
\newblock \emph{arXiv}, 2018.

\bibitem[Pumarola et~al.(2021)Pumarola, Corona, Pons-Moll, and Moreno-Noguer]{pumarola2021d}
Albert Pumarola, Enric Corona, Gerard Pons-Moll, and Francesc Moreno-Noguer.
\newblock D-nerf: Neural radiance fields for dynamic scenes.
\newblock In \emph{CVPR}, 2021.

\bibitem[Raoult et~al.(2017)Raoult, Reid-Anderson, Ferri, and Williamson]{raoult2017reliable}
Vincent Raoult, Sarah Reid-Anderson, Andreas Ferri, and Jane~E Williamson.
\newblock How reliable is structure from motion (sfm) over time and between observers? a case study using coral reef bommies.
\newblock \emph{Remote Sensing}, 2017.

\bibitem[Schonberger and Frahm(2016)]{schonberger2016structure}
Johannes~L Schonberger and Jan-Michael Frahm.
\newblock Structure-from-motion revisited.
\newblock In \emph{CVPR}, 2016.

\bibitem[Shao et~al.(2023)Shao, Zheng, Tu, Liu, Zhang, and Liu]{shao2023tensor4d}
Ruizhi Shao, Zerong Zheng, Hanzhang Tu, Boning Liu, Hongwen Zhang, and Yebin Liu.
\newblock Tensor4d: Efficient neural 4d decomposition for high-fidelity dynamic reconstruction and rendering.
\newblock In \emph{CVPR}, 2023.

\bibitem[Song et~al.(2023)Song, Chen, Li, Chen, Chen, Yuan, Xu, and Geiger]{song2023nerfplayer}
Liangchen Song, Anpei Chen, Zhong Li, Zhang Chen, Lele Chen, Junsong Yuan, Yi Xu, and Andreas Geiger.
\newblock Nerfplayer: A streamable dynamic scene representation with decomposed neural radiance fields.
\newblock \emph{IEEE Transactions on Visualization and Computer Graphics}, 2023.

\bibitem[Sudre et~al.(2017)Sudre, Li, Vercauteren, Ourselin, and Jorge~Cardoso]{10.1007/978-3-319-67558-9_28}
Carole~H. Sudre, Wenqi Li, Tom Vercauteren, Sebastien Ourselin, and M. Jorge~Cardoso.
\newblock Generalised dice overlap as a deep learning loss function for highly unbalanced segmentations.
\newblock In \emph{Deep Learning in Medical Image Analysis and Multimodal Learning for Clinical Decision Support}, 2017.

\bibitem[Tian et~al.(2023)Tian, Du, and Duan]{23iccv/tian_mononerf}
Fengrui Tian, Shaoyi Du, and Yueqi Duan.
\newblock {MonoNeRF}: Learning a generalizable dynamic radiance field from monocular videos.
\newblock In \emph{ICCV}, 2023.

\bibitem[Tretschk et~al.(2021)Tretschk, Tewari, Golyanik, Zollh{\"o}fer, Lassner, and Theobalt]{tretschk2021non}
Edgar Tretschk, Ayush Tewari, Vladislav Golyanik, Michael Zollh{\"o}fer, Christoph Lassner, and Christian Theobalt.
\newblock Non-rigid neural radiance fields: Reconstruction and novel view synthesis of a dynamic scene from monocular video.
\newblock In \emph{ICCV}, 2021.

\bibitem[Wang et~al.(2021)Wang, Wu, Xie, Chen, and Prisacariu]{nerfmm}
Zirui Wang, Shangzhe Wu, Weidi Xie, Min Chen, and Victor~Adrian Prisacariu.
\newblock Nerf-: Neural radiance fields without known camera parameters.
\newblock \emph{CoRR}, 2021.

\bibitem[Weng et~al.(2022)Weng, Curless, Srinivasan, Barron, and Kemelmacher-Shlizerman]{weng_humannerf_2022_cvpr}
Chung-Yi Weng, Brian Curless, Pratul~P. Srinivasan, Jonathan~T. Barron, and Ira Kemelmacher-Shlizerman.
\newblock Human{N}e{RF}: Free-viewpoint rendering of moving people from monocular video.
\newblock In \emph{CVPR}, 2022.

\bibitem[Wu et~al.()Wu, Yi, Fang, Xie, Zhang, Wei, Liu, Tian, and Wang]{Wu_2024_CVPR}
Guanjun Wu, Taoran Yi, Jiemin Fang, Lingxi Xie, Xiaopeng Zhang, Wei Wei, Wenyu Liu, Qi Tian, and Xinggang Wang.
\newblock 4d gaussian splatting for real-time dynamic scene rendering.
\newblock In \emph{CVPR}.

\bibitem[Yang et~al.(2022)Yang, Vo, Natalia, Ramanan, Andrea, and Hanbyul]{yang2022banmo}
Gengshan Yang, Minh Vo, Neverova Natalia, Deva Ramanan, Vedaldi Andrea, and Joo Hanbyul.
\newblock Banmo: Building animatable 3d neural models from many casual videos.
\newblock In \emph{CVPR}, 2022.

\bibitem[Yang et~al.(2023)Yang, Gao, Li, Gao, Wang, and Zheng]{yang2023track}
Jinyu Yang, Mingqi Gao, Zhe Li, Shang Gao, Fangjing Wang, and Feng Zheng.
\newblock Track anything: Segment anything meets videos.
\newblock \emph{arXiv}, 2023.

\bibitem[Yang et~al.(2024)Yang, Gao, Zhou, Jiao, Zhang, and Jin]{yang2023deformable3dgs}
Ziyi Yang, Xinyu Gao, Wen Zhou, Shaohui Jiao, Yuqing Zhang, and Xiaogang Jin.
\newblock Deformable 3d gaussians for high-fidelity monocular dynamic scene reconstruction.
\newblock In \emph{CVPR}, 2024.

\bibitem[Yoon et~al.(2020)Yoon, Kim, Gallo, Park, and Kautz]{yoon2020dynamic}
Jae~Shin Yoon, Kihwan Kim, Orazio Gallo, Hyun~Soo Park, and Jan Kautz.
\newblock Novel view synthesis of dynamic scenes with globally coherent depths from a monocular camera.
\newblock In \emph{CVPR}, 2020.

\bibitem[Zhang et~al.(2018)Zhang, Isola, Efros, Shechtman, and Wang]{zhang2018perceptual}
Richard Zhang, Phillip Isola, Alexei~A Efros, Eli Shechtman, and Oliver Wang.
\newblock The unreasonable effectiveness of deep features as a perceptual metric.
\newblock In \emph{CVPR}, 2018.

\end{thebibliography}
}
\clearpage
\setcounter{page}{1}
\maketitlesupplementary

\appendix

\section{Demo Videos}
We recommend that readers refer to our project page at \url{https://kaist-viclab.github.io/splinegs-site/}, which showcases extensive qualitative comparisons between our SplineGS and SOTA novel view synthesis methods~\cite{Wu_2024_CVPR, Li_STG_2024_CVPR, liu2023robust, yang2023deformable3dgs, lee2024fully, Gao-ICCV-DynNeRF}. Please note that since the provided project page for this supplementary material is \textit{offline}, and therefore, \textit{no modifications can be made after submission}; it is offered solely for the convenience of visualization. The project page features various demo videos, including comparisons for (i) novel view synthesis on NVIDIA~\cite{yoon2020dynamic}, (ii) novel view and time synthesis on NVIDIA~\cite{yoon2020dynamic}, (iii) novel view synthesis on DAVIS~\cite{ponttuset20182017davischallengevideo}, showcasing fixed views, spiral views, and zoomed-in/out views, (iv) dynamic 3D Gaussian trajectory visualization on DAVIS~\cite{ponttuset20182017davischallengevideo}, and (v) visualization of a toy example when we edit 3D positions of several control points.

\section{Additional Ablation Study for Motion-Adaptive Control Points Pruning (MACP)}
\label{sec:macp_epsilon_ablation}

\begin{figure}[ht]
    \centering
    \includegraphics[width=\linewidth,keepaspectratio]{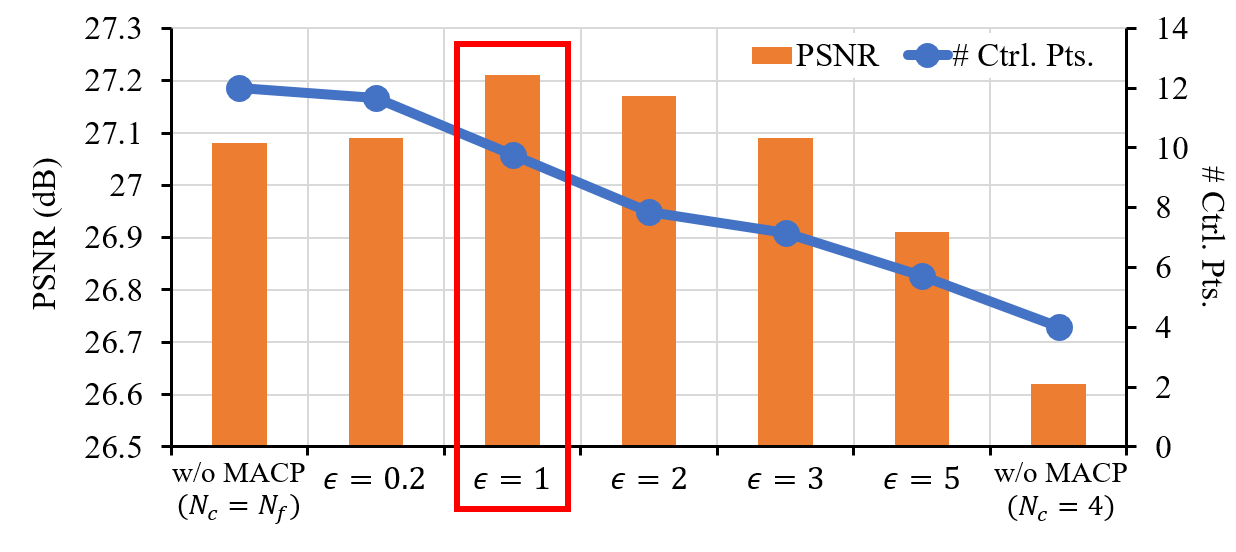}
    \caption{\textbf{Ablation study on MACP.} We conduct an ablation study of our Motion-Adaptive Control points Pruning (MACP) method for novel view synthesis on the NVIDIA dataset~\cite{yoon2020dynamic} by adjusting the pruning error threshold $\epsilon$. `PSNR (dB)' and `\# Ctrl. Pts.' denote the average PSNR value and the average number of control points for dynamic 3D Gaussians after training, computed across all scenes, respectively.}
    \label{fig:sup_macp}
\end{figure}

As described in Eq.~\ref{eq:2d_thershold_macp} of the main paper, we compute the error $E$ between $S(t, \textbf{P})$ and $S(t, \textbf{P}')$ by projecting the 3D points of each cubic Hermite spline function~\cite{ahlberg2016theory, de1978practical} over time into pixel space of all training cameras. This error is then used to update the new spline function. The 2D error measurement is particularly effective because it directly aligns with the image domain, where pixel-level accuracy is essential for precise spline function updates. To determine the updated spline function, we set the threshold value $\epsilon$ of the error $E$ in Eq.~\ref{eq:2d_thershold_macp} to 1.
To validate the rationale behind our setup, we conduct an ablation study for novel view synthesis on the NVIDIA dataset~\cite{yoon2020dynamic}, examining different MACP settings, including the ablated models without MACP (`w/o MACP ($N_c=4$)', `w/o MACP ($N_c=N_f$)' in Table~\ref{table:ablation_study}-(c)) and with MACP having variations in $\epsilon$ values.
For the variations in $\epsilon$ values, we select 0.2, 1, 2, 3, and 5.
Fig.~\ref{fig:sup_macp} presents the average PSNR values and the average number of control points for dynamic 3D Gaussians after training across all scenes.
As shown in Fig.~\ref{fig:sup_macp}, when $\epsilon$ is set to an excessively small value (`$\epsilon = 0.2$'), our MAS architecture fails to prune control points effectively, resulting in reduced efficiency. Conversely, when $\epsilon$ is too large (`$\epsilon = 5$'), the pruning becomes overly aggressive, resulting in an insufficient number of control points to accurately represent complex motion trajectories. This trade-off underscores the importance of selecting $\epsilon$ carefully to achieve a balance between efficiency and representation quality.

\section{Memory Footprint Comparison}
\label{sec:complexity_comparison}
To further highlight the efficiency of our SplineGS, we compared its memory footprint with other 3DGS-based methods \cite{Wu_2024_CVPR, yang2023deformable3dgs, lee2024fully, Li_STG_2024_CVPR}, as shown in Table~\ref{table:memory_footprint}. This comparison evaluates the average model storage requirements after optimization on the NVIDIA dataset~\cite{yoon2020dynamic}. The storage requirements of 3DGS-based methods depend on the number of 3D Gaussians, which is determined by their hyperparameters. For consistency, we use the same hyperparameter settings for the 3DGS-based methods~\cite{Wu_2024_CVPR, yang2023deformable3dgs, lee2024fully, Li_STG_2024_CVPR} as those specified in their original implementations. Ex4DGS \cite{lee2024fully} requires the largest memory footprint, attributed to its method of explicit keyframe dynamic 3D Gaussian fusion. In contrast, our SplineGS, which achieves state-of-the-art (SOTA) rendering quality as shown in Table~\ref{table:nvidia_quantitative}, utilizes only about one-tenth of the memory footprint required by Ex4DGS~\cite{lee2024fully}, thanks to our efficient MAS representation and the MACP method.

\begin{table}[h]
\begin{center}
\scalebox{0.7}{
\begin{tabular}{ l | c | c }
\toprule
\multicolumn{1}{c|}{Method} & Memory footprint (MB) $\downarrow$ & \# Gaussian (K)  \\  
\bottomrule
\hline\noalign{\smallskip}
4DGS (CVPR'24)~\cite{Wu_2024_CVPR}    & 50 & 136  \\
D3DGS (CVPR'24)~\cite{yang2023deformable3dgs} & 92  &  382 \\
Ex4DGS (NeurIPS'24)~\cite{lee2024fully}  & 256    & 436  \\
STGS (CVPR'24)~\cite{Li_STG_2024_CVPR}  & \best{19} &  128  \\
\midrule
\textbf{SplineGS (Ours)} & \second{26} & 183  \\
\bottomrule
\end{tabular}
}
\vspace{-0.2cm}
\caption{\textbf{Memory footprint comparison results.} `Memory footprint (MB)' refers to the memory size of each trained model, while `\# Gaussian (K)' represents the total number of 3D Gaussians after training.}
\label{table:memory_footprint}
\vspace{-0.4cm}
\end{center}
\end{table}

\section{Dynamic 3D Gaussian Trajectory Visualization}
\label{sec:additional_motion_tracking}
Please note that the term \textit{motion tracking} in our main paper (Fig.~\ref{fig:tracking}), also referred to as dynamic 3D Gaussian trajectory visualization in 2D space, differs from the term \textit{tracking} used in \textit{2D Tracking} methods such as~\cite{karaev2023cotracker}, which aim to find 2D pixel correspondences among \textit{given} video frames. Our SplineGS leverages spline-based motion modeling to directly capture the deformation of each dynamic 3D Gaussian along the temporal axis, enabling the rendering of target novel views. For 2D visualization of the 3D motion of each dynamic 3D Gaussian, which is referred to as \textit{motion tracking} in our main paper, we project its trajectory onto the 2D pixel space of the novel views. We compute a \textit{rasterized} 2D track $\bm{\mathcal{T}}^G = \{\bm{\varphi}^G_{t'} | \bm{\varphi}^G_{t'} \in \mathbb{R}^2 \}_{t' \in [t_1, t_2]}$ over the specified time interval $[t_1, t_2]$ as the Gaussians' trajectories visualization shown in Fig.~\ref{fig:tracking} of the main paper. For this motion tracking rasterization, we compute the projected pixel coordinates at time $t'$ for each 3D Gaussian using the camera pose $[\bm{R}^*|\bm{T}^*]$ of the target novel view as $\pi_{\hat{\bm{K}}}(\bm{R}^*S(t', \textbf{P}) + \bm{T}^*)$. Then, we compute $\bm{\varphi}^G_{t'}$ by replacing the color $\bm{c}_i$ in Eq.~\ref{eq:alpha_blending} with the projected pixel coordinate as
\begin{equation}
    \bm{\varphi}^G_{t'} = \textstyle \sum_{i\in\mathcal{N}} \pi_{\hat{\bm{K}}}(\bm{R}^*S_i(t', \textbf{P}) + \bm{T}^*) \alpha^\text{dy}_i\prod^{i-1}_{j=1}(1-\alpha^\text{dy}_j),
\end{equation}
where $\alpha^\text{dy}_i$ denotes the density of the $i^\text{th}$ dynamic 3D Gaussian. 
\begin{figure}[ht]
    \centering
    \includegraphics[width=\linewidth,keepaspectratio]{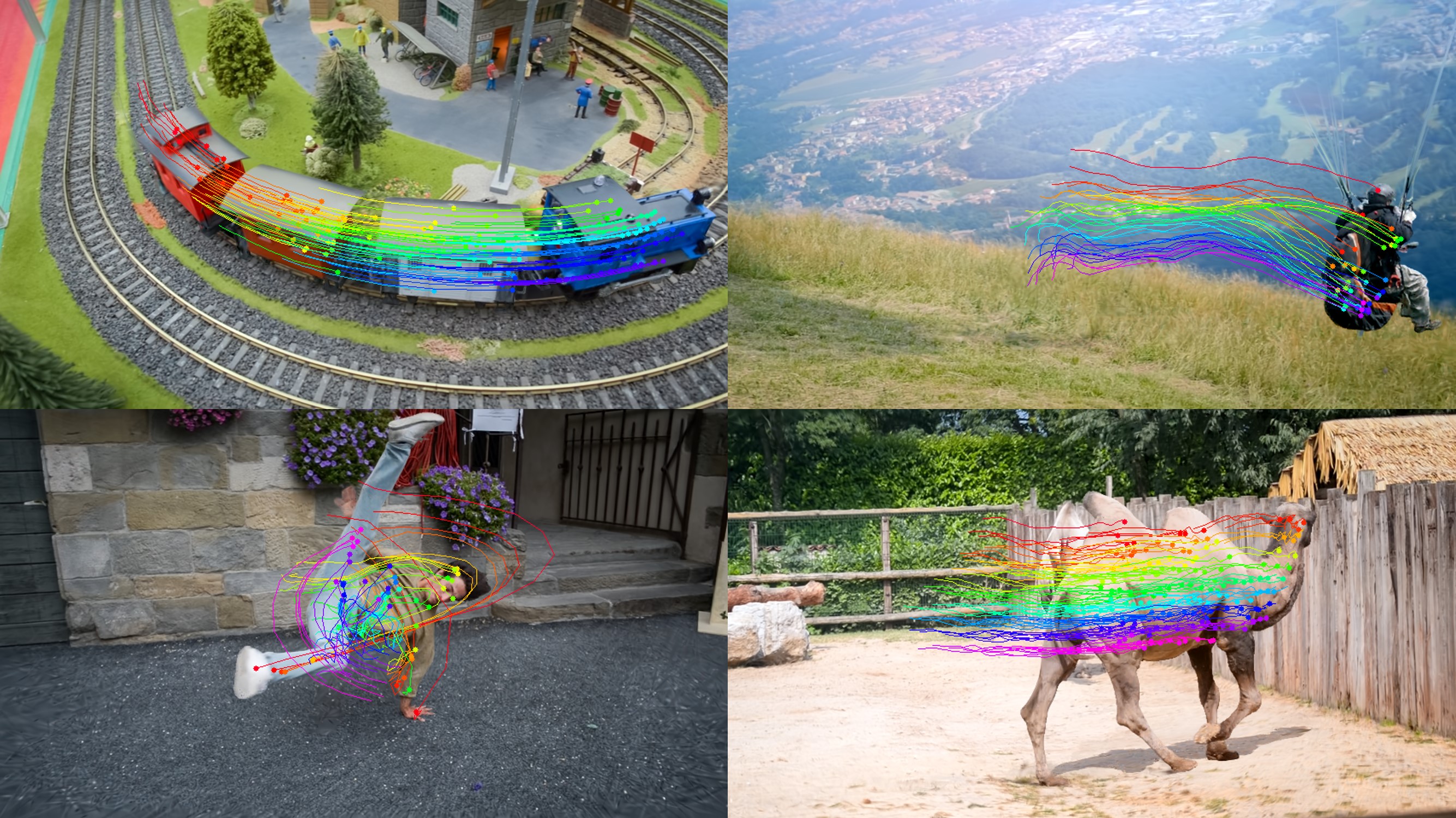}
    \caption{\textbf{Visual results of dynamic 3D Gaussian trajectory projected to novel views for our SplineGS.}}
    \label{fig:tracking_ours}
\end{figure}

\begin{figure}[ht]
    \centering
    \includegraphics[width=\linewidth,keepaspectratio]{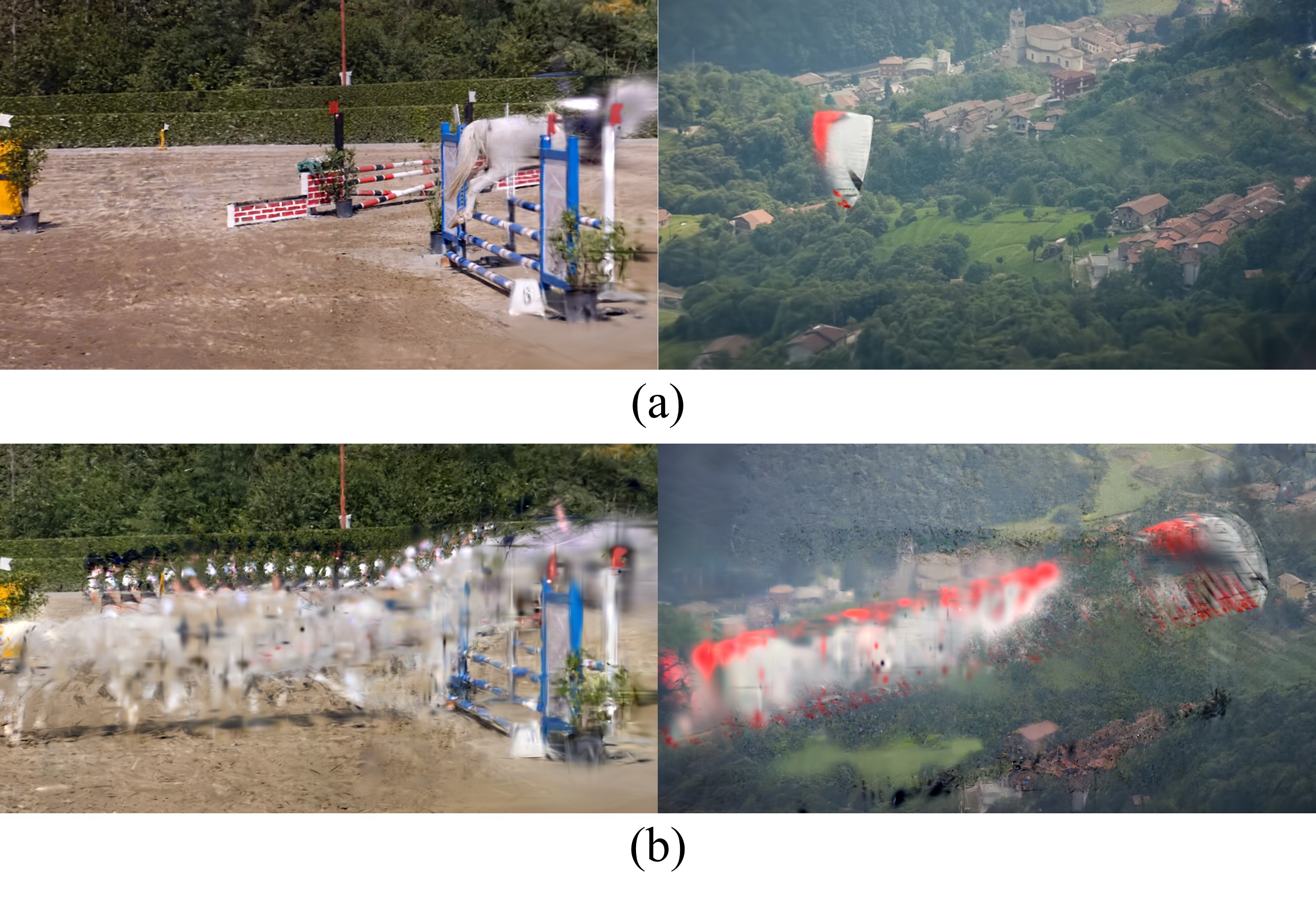}
    \caption{\textbf{Visual results of novel view synthesis at a specific time using the same STGS~\cite{Li_STG_2024_CVPR} models after optimization with (a) their original time-varying opacity and (b) time-independent spatial opacity, respectively.} Please note that we use their original time-varying opacity during training.}
    \label{fig:opacity_stgs}
\end{figure}

As shown in Fig.~\ref{fig:tracking} of the main paper, D3DGS~\cite{yang2023deformable3dgs} fails to reconstruct dynamic regions. STGS~\cite{Li_STG_2024_CVPR} renders dynamic regions more effectively than D3DGS~\cite{yang2023deformable3dgs}, but it still produces poor visualizations of 3D Gaussian trajectories. 
In the original STGS~\cite{Li_STG_2024_CVPR} paper, they propose the temporal opacity $\sigma_i(t)$ as 
\begin{equation}
    \sigma_i(t) = \sigma^s_i \exp(-s_i^\tau|t-\mu_i^\tau|^2),
\end{equation}
where $\mu_i^\tau$ is the temporal center, $s_i^\tau$ is the temporal scaling factor and $\sigma^s_i$ is the time-independent spatial opacity. 
To further investigate the motion tracking results of STGS~\cite{Li_STG_2024_CVPR}, we render novel views for STGS~\cite{Li_STG_2024_CVPR} after training by setting the opacity of each 3D Gaussian with (a) its original temporal opacity $\sigma_i(t)$ and (b) the fixed value of time-independent spatial opacity $\sigma^s_i$, as shown in Fig.~\ref{fig:opacity_stgs}.

We observe that when the opacity of each 3D Gaussian is set to a time-independent value, the rendered novel view synthesis results show multiple instances of the same moving objects (e.g. a horse or a parachute) appearing simultaneously, as illustrated in Fig.~\ref{fig:opacity_stgs}-(b). This observation suggests that, to represent a moving object across time, STGS~\cite{Li_STG_2024_CVPR} may adjust the opacities of \textit{different} sets of 3D Gaussians through their temporal opacities $\sigma_i(t)$, rather than deforming the spatial 3D positions of a \textit{single} set of 3D Gaussians along the temporal axis. While this approach can produce dynamic rendering results, it may not allow for the direct extraction of 3D Gaussian trajectories along the temporal axis. In contrast, our SplineGS with MAS directly models the motion trajectories of dynamic 3D Gaussians, enabling the extraction of more reasonable 3D trajectories, as shown in Fig.~\ref{fig:tracking_ours}.

\section{Additional Details for Methodology}
\label{sec:implement_details}
\noindent \textbf{Camera Intrinsic.} To predict the shared camera intrinsics for our camera parameter estimation, we adopt a pinhole camera model which is widely used in COLMAP-free novel view synthesis methods~\cite{liu2023robust, nerfmm, meng2021gnerf, camp} as
\begin{equation}
    \bm{K} = \begin{bmatrix}
        f_x & s & c_x \\ 0 & f_y & c_y \\ 0&0&1
    \end{bmatrix},
\end{equation}
where $s=0$ represents the skewness of the camera, while $c_x$ and $c_y$ denote the coordinates of the principal point in pixels. Without loss of generality, we assume that $f_x = f_y = f$, indicating equal focal lengths in both directions, and set $c_x$ and $c_y$ to half the width and height of the video frame, respectively.

\noindent \textbf{Time-dependent Rotation and Scale.} As described in Sec.~\ref{sec:preliminary} of the main paper, we model the rotation and scale of dynamic 3D Gaussians as time-dependent functions. For the rotation, we adopt a polynomial function inspired by STGS~\cite{Li_STG_2024_CVPR}, defined as
\begin{equation}
    \bm{q}_i(t) = \bm{q}^0_i + \textstyle \sum^{n_q}_{k=1}\Delta \bm{q}_{i,k}t^k,
\end{equation}
where $\bm{q}^0_i$ is a time-independent base quaternion of the $i^\text{th}$ dynamic 3D Gaussian and $\Delta \bm{q}_{i,k}$ is an offset quaternion of the $k^\text{th}$-order term of $i^\text{th}$ dynamic 3D Gaussian, both of which are learnable parameters. we set $n_q = 1$. This ensures a simple yet effective representation of time-dependent rotations~\cite{Li_STG_2024_CVPR}.
For the scale, inspired by DynIBaR~\cite{li2023dynibar}, we leverage the Discrete Cosine Transform (DCT) to capture the continuously varying scale of each dynamic 3D Gaussian. The scale function is expressed as
\begin{equation}
\begin{aligned}
    &\bm{s}_i(t) = \bm{s}^0_i + \Delta\bm{s}_i(t), \\
    &\Delta\bm{s}_i(t) = \sqrt{2/N_f} \textstyle \sum^K_{k=1}\zeta_{i,k}\cos\left(\frac{\pi}{2N_f}(2t+1)k\right),
\end{aligned}
\end{equation}
where $\bm{s}^0_i$ is a time-independent base scale vector of the $i^\text{th}$ dynamic 3D Gaussian and $\zeta_{i,k} \in \mathbb{R}^3$ represents the $k^\text{th}$ coefficient of the $i^\text{th}$ dynamic 3D Gaussian, both of which are learnable parameters. Here, $K = 10$ controls the number of frequency components used in the DCT, allowing flexible yet compact modeling of temporal scale variations.

\section{Limitation}
\begin{figure}[ht]
    \centering
    \includegraphics[width=\linewidth,keepaspectratio]{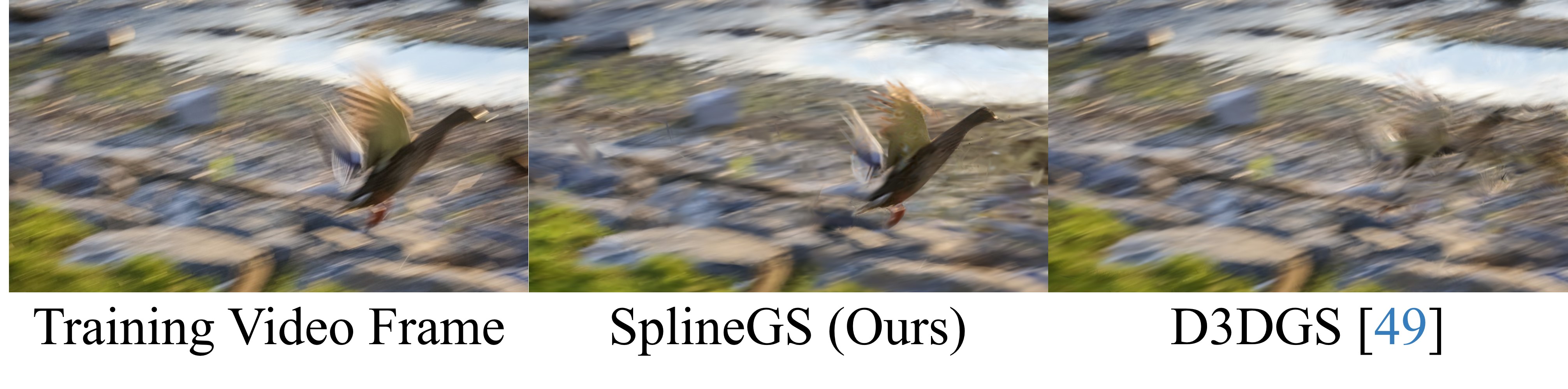}
    \caption{\textbf{Limitations of our SplineGS.} When the training video frame contains blurriness, our model cannot effectively reconstruct sharp renderings due to the absence of a deblurring method.}
    \label{fig:limitation}
\end{figure}

In-the-wild videos often exhibit significant and rapid camera and object movements, resulting in blurry input frames. This blurriness subsequently degrades the quality of the rendered novel views. As shown in Fig.~\ref{fig:limitation}, the methods solely designed for dynamic scene reconstruction may overfit to the blurry training frames. A straightforward solution is to employ state-of-the-art 2D deblurring methods to enhance the quality of input frames. Additionally, in future research, we plan to integrate a deblurring approach directly into the reconstruction pipeline. This integration could establish a joint deblurring and rendering optimization framework, addressing low-quality issues and enhancing the final rendered outputs without requiring separate preprocessing.

\section{Additional Qualitative Results}
\label{sec:additional_qualitative}

\subsection{Novel View Synthesis on NVIDIA}
\begin{figure*}[t]
    \centering
    \includegraphics[width=\linewidth,keepaspectratio]{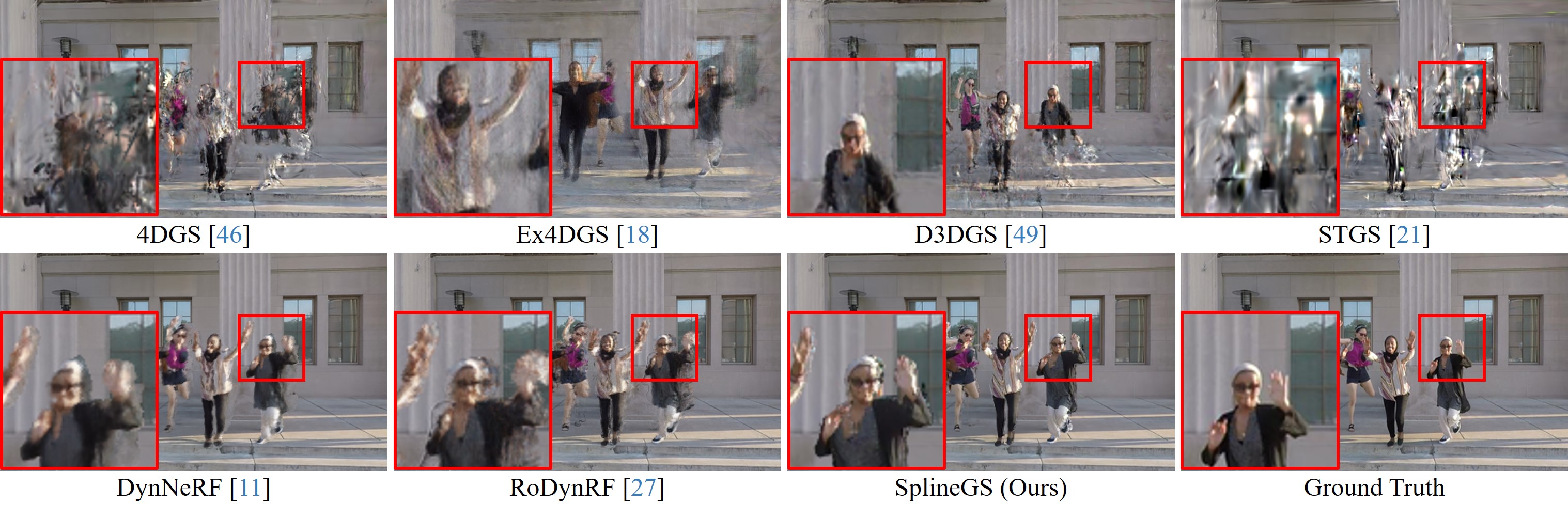}
    \caption{\textbf{Visual comparisons for novel view synthesis on the \textit{Jumping} scene from the NVIDIA dataset.}}
    \label{fig:qualitative_supple_nvidia_jumping}
\end{figure*}

\begin{figure*}[t]
    \centering
    \includegraphics[width=\linewidth,keepaspectratio]{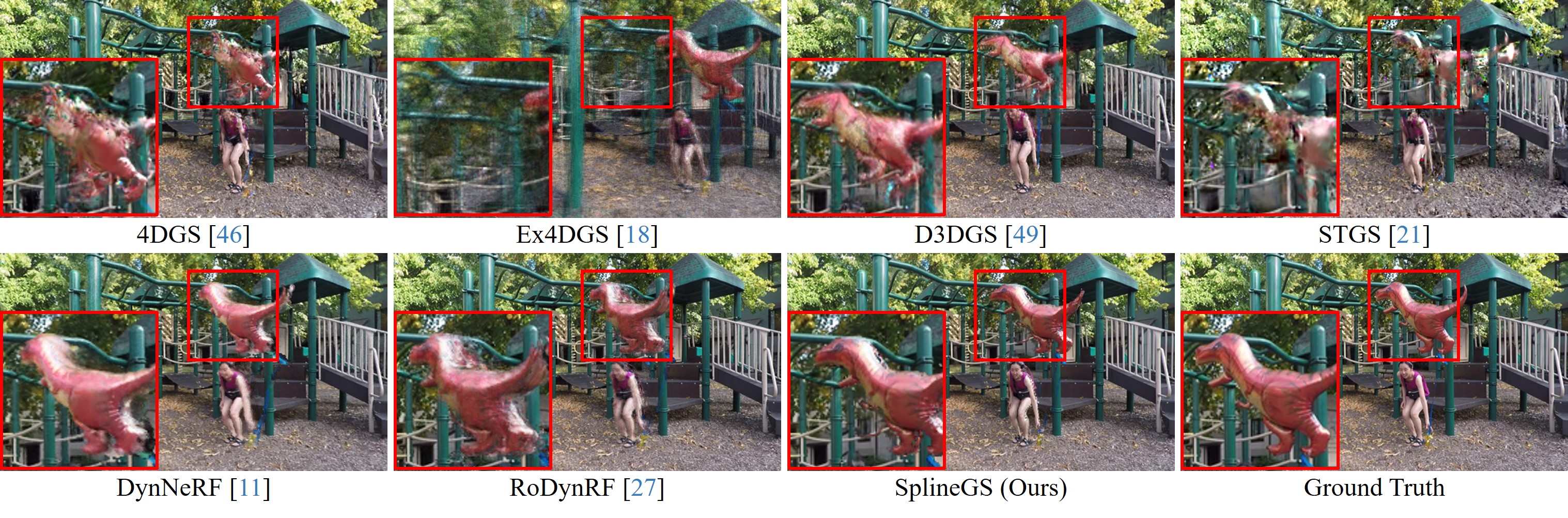}
    \caption{\textbf{Visual comparisons for novel view synthesis on the \textit{Playground} scene from the NVIDIA dataset.}}
    \label{fig:qualitative_supple_nvidia_playground}
\end{figure*}

\begin{figure*}[t]
    \centering
    \includegraphics[width=\linewidth,keepaspectratio]{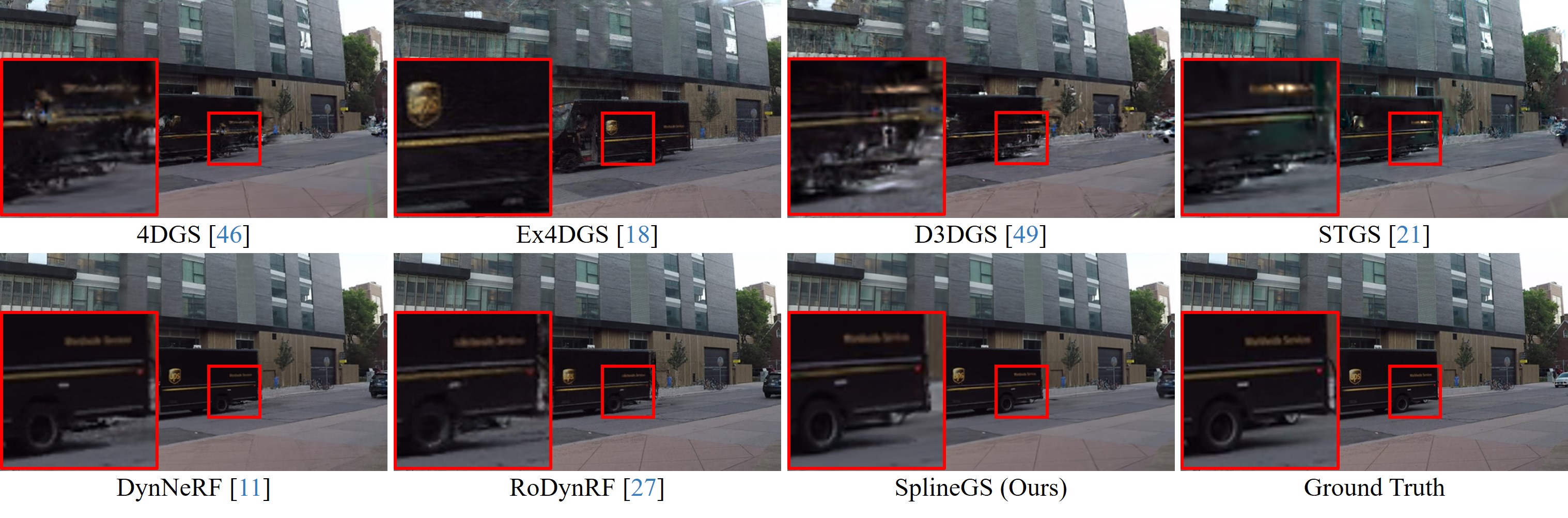}
    \caption{\textbf{Visual comparisons for novel view synthesis on the \textit{Truck} scene from the NVIDIA dataset.}}
    \label{fig:qualitative_supple_nvidia_truck}
\end{figure*}

Figs. \ref{fig:qualitative_supple_nvidia_jumping}, \ref{fig:qualitative_supple_nvidia_playground}, and \ref{fig:qualitative_supple_nvidia_truck} present additional visual comparisons for novel view synthesis on the NVIDIA dataset~\cite{yoon2020dynamic}.

\subsection{Novel View and Time Synthesis on NVIDIA}
\begin{figure*}[t]
    \centering
    \includegraphics[width=0.75\linewidth,keepaspectratio]{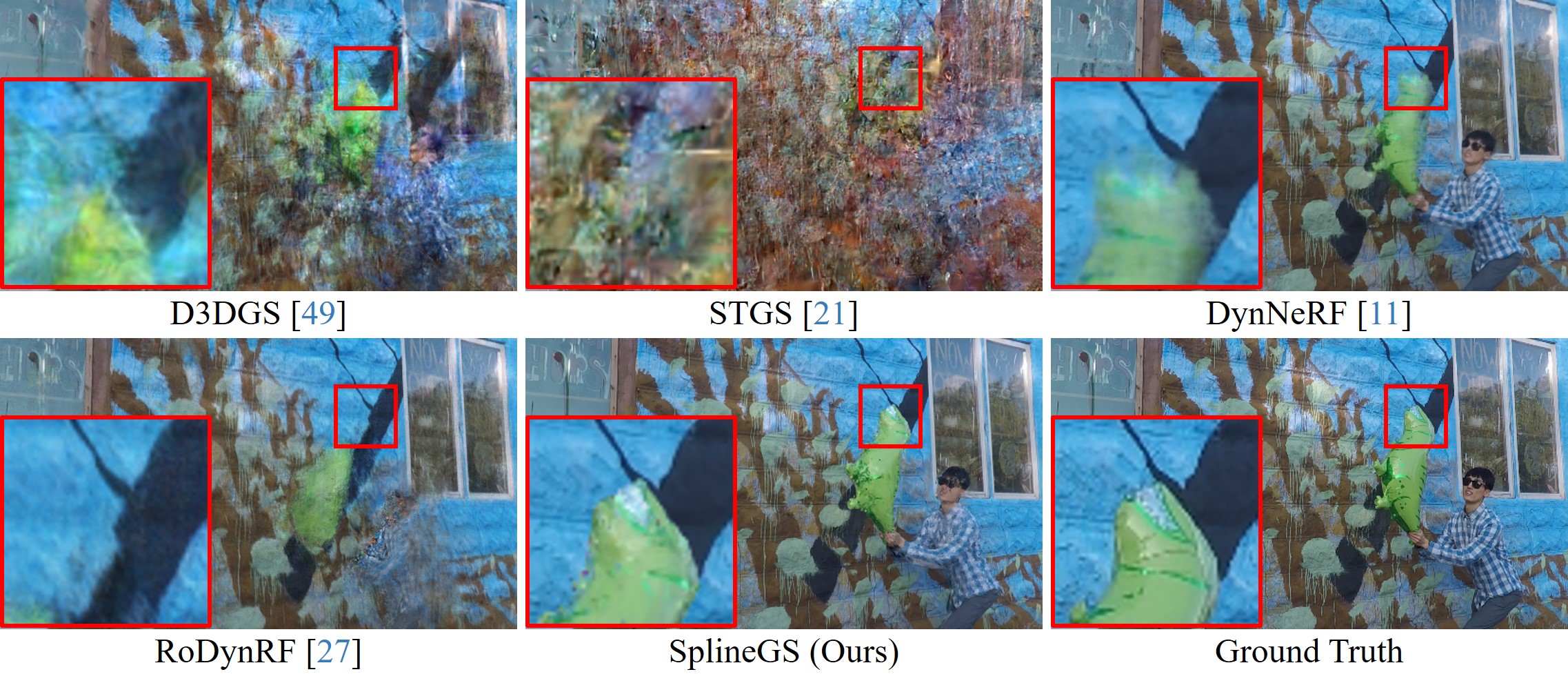}
    \caption{\textbf{Visual comparisons for novel view and time synthesis on the \textit{Balloon2} scene from the NVIDIA dataset.}}
    \label{fig:qualitative_supple_nvidia_nvts_balloon2}
\end{figure*}

\begin{figure*}[t]
    \centering
    \includegraphics[width=0.75\linewidth,keepaspectratio]{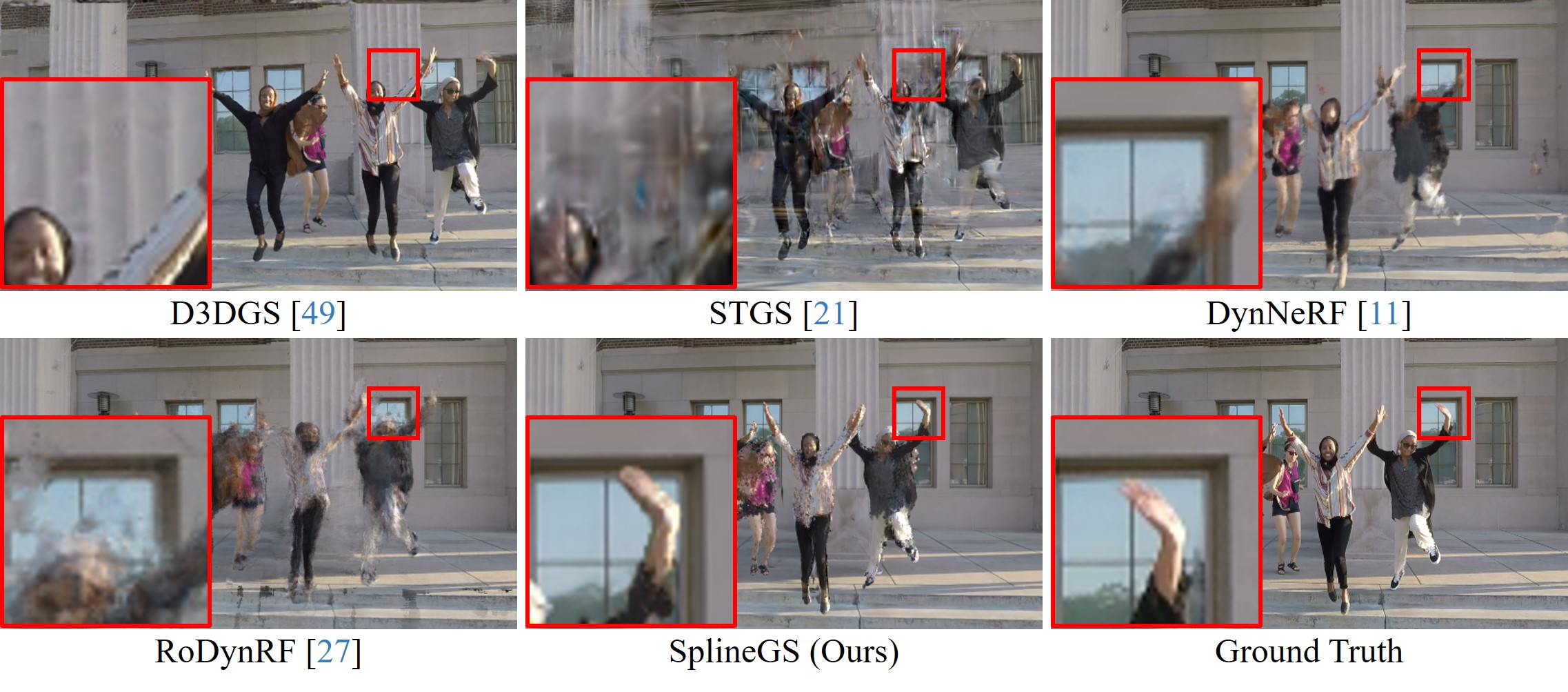}
    \caption{\textbf{Visual comparisons for novel view and time synthesis on the \textit{Jumping} scene from the NVIDIA dataset.}}
    \label{fig:qualitative_supple_nvidia_nvts_jumping}
\end{figure*}

\begin{figure*}[t]
    \centering
    \includegraphics[width=0.75\linewidth,keepaspectratio]{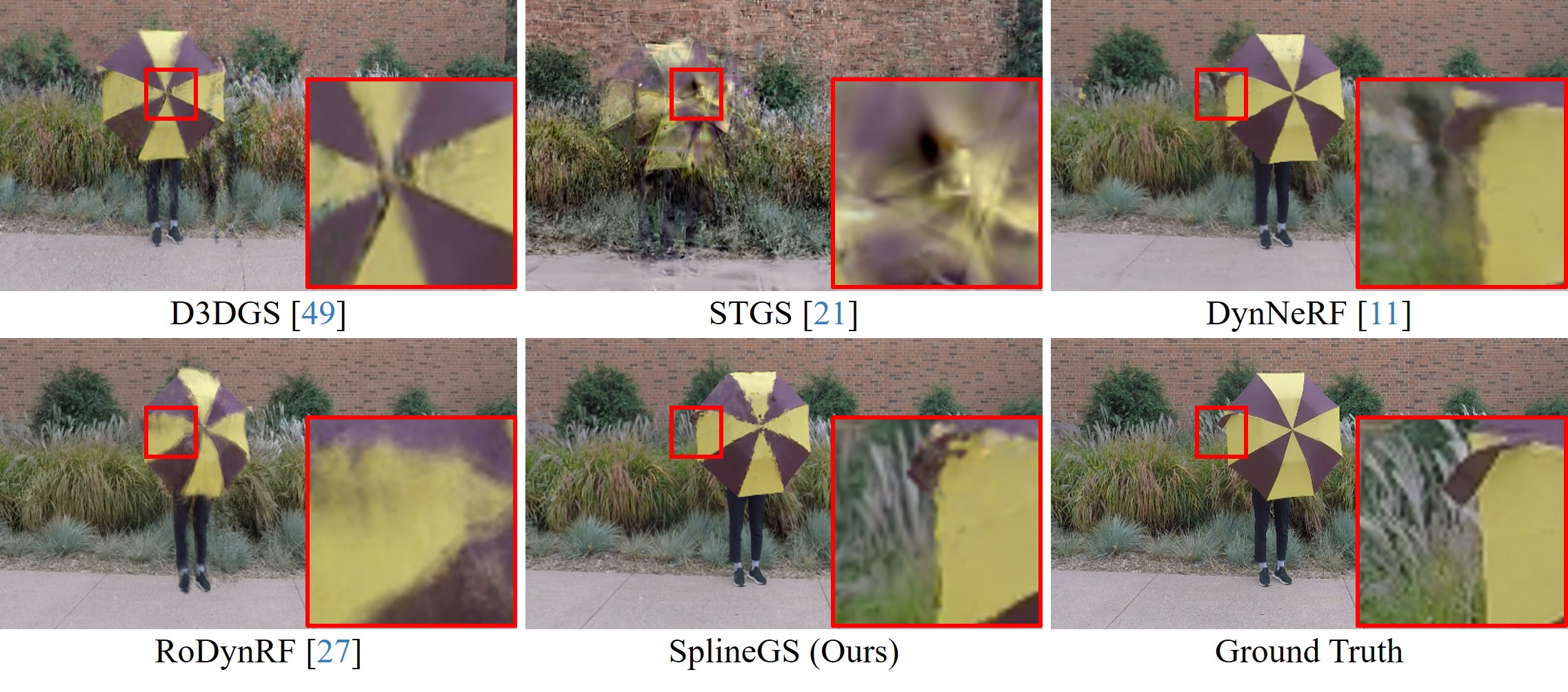}
    \caption{\textbf{Visual comparisons for novel view and time synthesis on the \textit{Umbrella} scene from the NVIDIA dataset.}}
    \label{fig:qualitative_supple_nvidia_nvts_umbrella}
\end{figure*}

Figs. \ref{fig:qualitative_supple_nvidia_nvts_balloon2}, \ref{fig:qualitative_supple_nvidia_nvts_jumping}, and \ref{fig:qualitative_supple_nvidia_nvts_umbrella} present additional visual comparisons for novel view and time synthesis on the NVIDIA dataset~\cite{yoon2020dynamic}.

\subsection{Novel View Synthesis on DAVIS}
\begin{figure*}[t]
    \centering
    \includegraphics[width=0.7\linewidth,keepaspectratio]{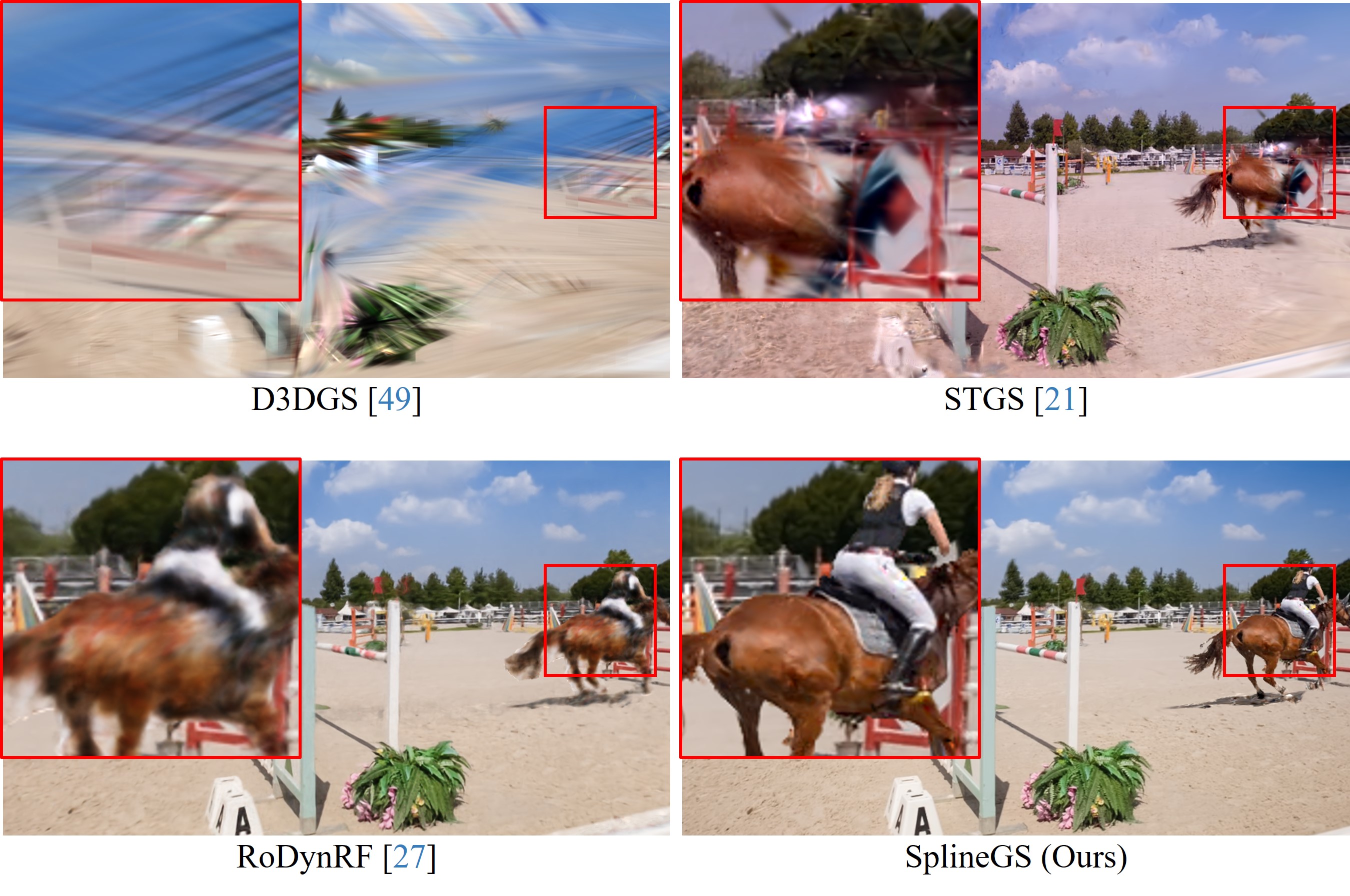}
    \caption{\textbf{Visual comparisons for novel view synthesis on the \textit{Horsejump-high} scene from the DAVIS dataset.}}
    \label{fig:qualitative_supple_davis_horsejump-high}
\end{figure*}

\begin{figure*}[t]
    \centering
    \includegraphics[width=0.7\linewidth,keepaspectratio]{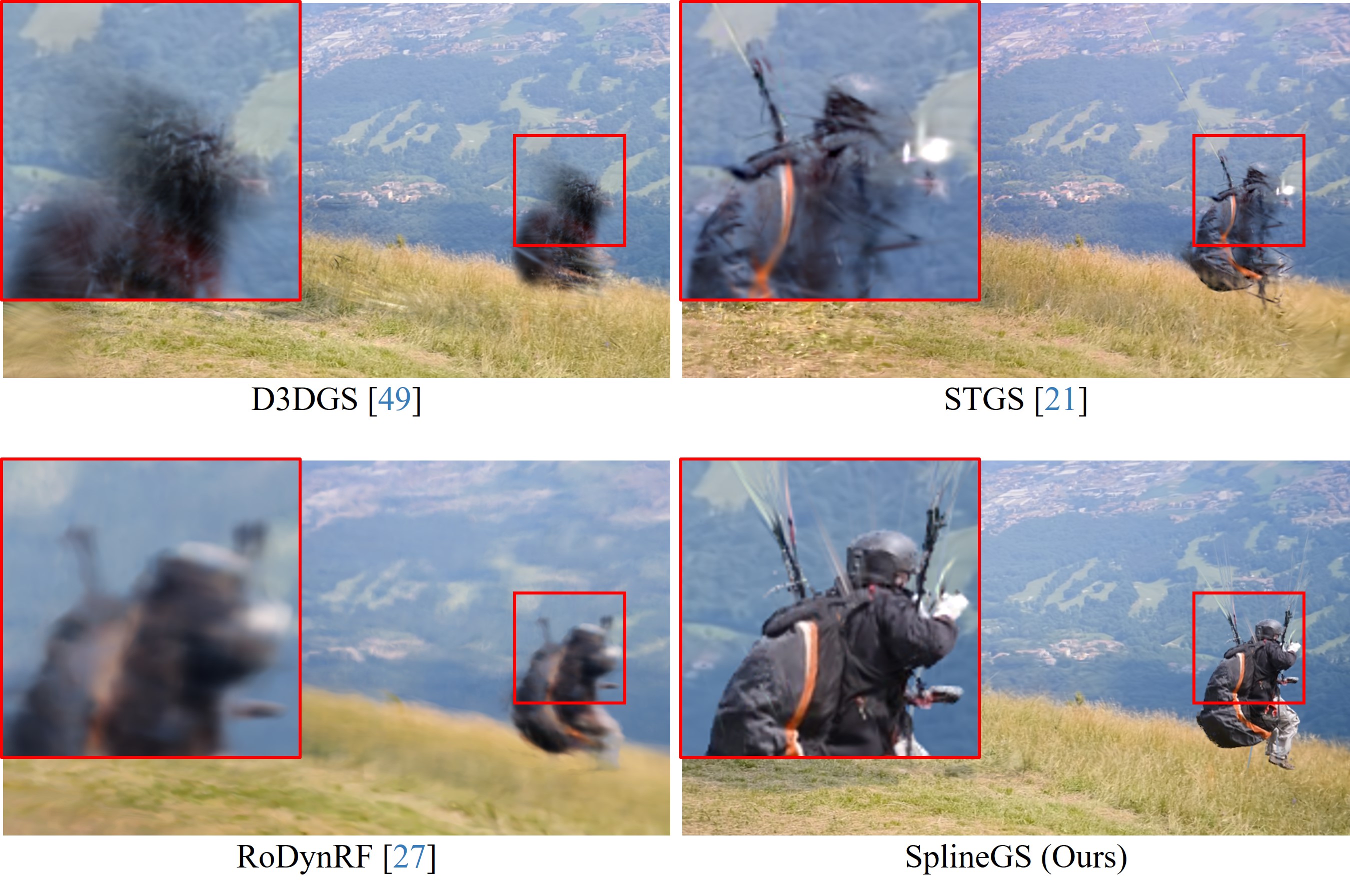}
    \caption{\textbf{Visual comparisons for novel view synthesis on the \textit{Paragliding-launch} scene from the DAVIS dataset.}}
    \label{fig:qualitative_supple_davis_paragliding-launch}
\end{figure*}

Figs. \ref{fig:qualitative_supple_davis_horsejump-high} and \ref{fig:qualitative_supple_davis_paragliding-launch} present additional visual comparisons for novel view synthesis on the DAVIS dataset~\cite{ponttuset20182017davischallengevideo}.

\clearpage

\clearpage

\clearpage

\end{document}